\documentclass{article}
\usepackage{preprint,times}
\usepackage{float}

\usepackage[utf8]{inputenc}
\usepackage[T1]{fontenc}
\usepackage{hyperref}
\hypersetup{colorlinks=true,citecolor=blue,linkcolor=blue,urlcolor=blue}
\usepackage{url}
\usepackage{xurl}
\usepackage{booktabs}
\usepackage{array}
\usepackage{amsfonts}
\usepackage{nicefrac}
\usepackage{microtype}
\usepackage{xcolor}
\definecolor{tsdiffcolor}{HTML}{1F77B4}
\definecolor{controltscolor}{HTML}{D62728}
\usepackage{graphicx}
\usepackage{subcaption}
\usepackage{tikz}
\usetikzlibrary{arrows.meta, positioning, calc, fit, decorations.markings}
\usepackage{amsmath}
\usepackage{amssymb}
\usepackage{mathtools}
\usepackage{amsthm}
\usepackage{comment}
\usepackage{pgfplots}
\pgfplotsset{compat=1.18}
\usepackage[capitalize,noabbrev]{cleveref}
\usepackage{algorithm}
\usepackage{algorithmic}
\usepackage{placeins}
\usepackage{multirow}
\usepackage{wrapfig}

\graphicspath{{./}}

\theoremstyle{plain}

\theoremstyle{definition}

\theoremstyle{remark}

\usepackage[disable,textsize=tiny]{todonotes}

\iclrfinalcopy

\title{Quantile-Guided Diffusion for Probabilistic Time-Series Forecasting}

\author{%
    Shadi Omari$^{*}$, Bashar Khoury$^{*}$ \& Assaf Schuster \\
    Department of Computer Science \\
    Technion -- Israel Institute of Technology \\
    \texttt{\{shadi-omari,bashar.kh\}@cs.technion.ac.il}, \texttt{assaf@technion.ac.il}
}

\begin{document}

\maketitle
{\renewcommand{\thefootnote}{\fnsymbol{footnote}}\footnotetext[1]{Equal contribution.}}
\lhead{Preprint}

\begin{abstract}
Quantile forecasts summarize uncertainty at each future time step, while diffusion forecasters represent it through sampled future trajectories. We study how predicted quantiles can serve as distributional controls for diffusion forecasting. Specifically, we investigate how the mechanism of quantile integration affects its effectiveness in guiding a pretrained diffusion forecaster. We find that residual control throughout the denoising network yields the most consistent improvements. We instantiate this approach as \emph{Control-TS}, which learns a quantile-control pathway while keeping the pretrained backbone frozen. Across eight benchmarks, Control-TS consistently improves the backbone’s CRPS, achieves competitive performance against state-of-the-art forecasting models, and retains its gains in capacity-matched comparisons. Further analyses examine how these gains depend on quantile granularity, predictor choice, and forecast quality. Finally, we introduce the \emph{Conditional Linear Predictive Score (CLPS)}, which complements distributional metrics by evaluating the downstream forecasting utility of generated futures. Together, these findings support predicted quantiles as an effective interface between explicit uncertainty estimates and diffusion-based trajectory generation.
\end{abstract}

\section{Introduction}
Time-series forecasting supports decision-making in domains such as healthcare, retail, renewable energy, finance, and climate \citep{healthcarets, MUDELSEE2019310, SEZER2020106181, articleretailts, articlerenewableenergy}. Point forecasts alone are often insufficient: decisions also depend on the range and likelihood of possible outcomes \citep{hyndman2021fpp3}. Probabilistic forecasting addresses this need by characterizing a distribution over possible future outcomes.

Predictive distributions can be represented in various ways. Quantile forecasts provide explicit, horizon-wise summaries of predictive location, dispersion, and asymmetry \citep{salinas2020deepar,lim2021tft}, while diffusion forecasters represent uncertainty through variation across sampled future trajectories \citep{bilos2023,kollovieh2023tsdiff}.

Forecasters trained through quantile regression or autoregressive likelihood estimation may capture predictive patterns that complement those learned through diffusion denoising. Their quantile forecasts could make this predictive information available throughout the denoising process. However, these quantiles are imperfect and do not specify dependence across future time steps. The challenge is to exploit these marginal summaries while leaving the diffusion model responsible for generating complete future trajectories. We therefore ask: \emph{Can externally predicted quantile trajectories provide effective guidance for diffusion forecasting, and if so, how should they be incorporated?}

\begin{figure}[t]
  \centering
\includegraphics[width=0.83\textwidth]{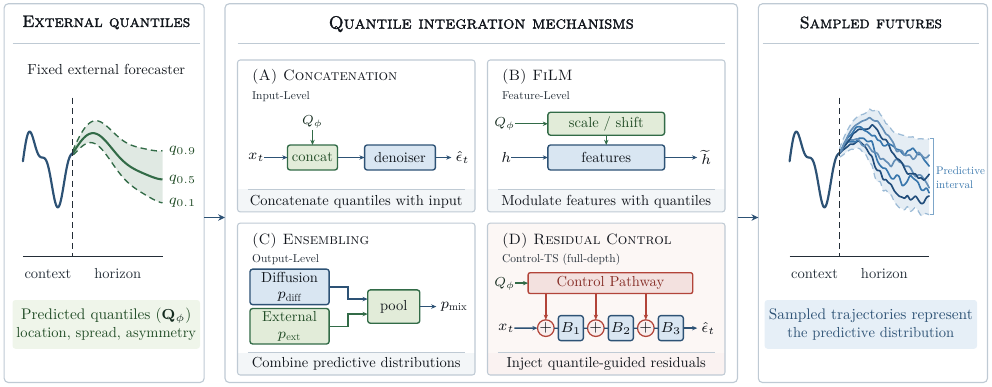}
  \caption{\textbf{Integrating predicted quantiles with diffusion forecasting.} (A) concatenates, (B) FiLM, (C) ensembles. (D) adds residuals before the first 1, 2, or all blocks (Control-TS). All improve backbone CRPS on some datasets; full-depth Control-TS gives the most consistent gains on all eight.}
  \label{fig:distributional_control}
\end{figure}

We study this question through \emph{distributional control}: using explicit summaries of a predictive distribution to guide diffusion forecasting. We use TSDiff \citep{kollovieh2023tsdiff} as our primary experimental backbone since it separates joint sequence modeling from forecast conditioning, enabling quantile control of a frozen pretrained denoiser. Starting from the same pretrained backbone and quantile forecasts, we investigate how output-level ensembling and input-, feature-, and residual-level conditioning affect forecasting performance, as summarized in Figure~\ref{fig:distributional_control}. Notably, every evaluated mechanism improves CRPS on some datasets, demonstrating the promise of external predictive information across distinct interfaces. Among these, full-depth residual control, instantiated as \emph{Control-TS}, is the only design to yield improvements across all eight benchmarks. This consistency motivates our focus on Control-TS in the subsequent analyses.

Control-TS converts predicted quantile trajectories into learned residual guidance for diffusion forecasting. It uses an auxiliary trainable controller to adapt a frozen denoiser, building on prior work on residual adapters, side-tuning, and spatial conditioning of image diffusion models (\citealp{rebuffi2018efficient}; \citealp{zhang2020sidetuning}; \citealp{controlnet}). The controller processes horizon-aligned quantiles and the diffusion timestep, producing residual corrections throughout the denoising network. To isolate the quantiles' contribution, we also evaluate observed history as an alternative control input. Additionally, we examine how quantile granularity, predictor choice, and degraded or misaligned controls affect guidance effectiveness. Analyses of controller design and computational cost characterize the resulting accuracy-cost tradeoff.

Across eight forecasting benchmarks, Control-TS improves CRPS over its diffusion backbone on every dataset, with reductions of up to approximately 35\%, and achieves the best or tied-best mean CRPS among all evaluated forecasters on five benchmarks. It also outperforms a capacity-matched diffusion backbone on six datasets and TSDiff's training-free refinement of external forecasts on seven, supporting benefits beyond increased capacity or access to external forecasts alone. Experiments with CSDI \citep{tashiro2021csdi} and MG-TSD \citep{fan2024mgtsd} further demonstrate the applicability of quantile control to other diffusion forecasters, with gains varying across backbones and datasets.

To complement distributional evaluation, we ask whether generated futures can serve as useful training targets for downstream forecasting. Horizon-wise CRPS measures marginal predictive accuracy \citep{gneiting2007scoring,hersbach2000crps}, but does not directly assess this utility. We introduce the \emph{Conditional Linear Predictive Score (CLPS)}, a Train-on-Hybrid, Test-on-Real (THTR) criterion \citep{articletstr} that trains a linear predictor on real contexts paired with generated futures and evaluates it on held-out real data. The shared linear protocol is computationally inexpensive, with deterministic fitting for fixed data. Lower scores indicate greater downstream utility under this protocol. Control-TS improves CLPS over its backbone on six of eight benchmarks.

\par\medskip
\noindent\textbf{Contributions.}\footnote{The code will be released upon publication.}
\begin{enumerate}
    \item \textbf{Quantile-guided diffusion forecasting.} We establish predicted quantile trajectories as effective distributional controls through a systematic study of interaction mechanisms, quantile granularity, predictor choice, and control quality. Full-depth residual control is the most consistently effective interface on our primary backbone.

    \item \textbf{Control-TS and empirical validation.} We instantiate this interface as Control-TS, improving the primary backbone on all eight benchmarks, outperforming a capacity-matched diffusion baseline on six, and achieving best or tied-best overall CRPS on five. Further experiments assess benefits across backbones, controller design, and computational tradeoffs.

    \item \textbf{Complementary predictive evaluation.} We introduce CLPS to assess generated futures' forecasting utility via a shared linear evaluator with inexpensive, deterministic fitting for fixed data. Control-TS outperforms its diffusion backbone on six of eight benchmarks.
\end{enumerate}

\section{Quantile Control for Diffusion Forecasting}
\label{sec:distributional-control}
\subsection{Quantile forecasts as distributional controls}
\label{subsec:distributional-control}

We consider univariate probabilistic forecasting. Let $\mathbf{y}_{\mathrm{ctx}}\in\mathbb{R}^{L_{\mathrm{ctx}}}$ denote the observed context and $\mathbf{Y}_{\mathrm{tar}}=(Y_1,\ldots,Y_H)$ the future trajectory, where $H=L_{\mathrm{tar}}$. Our objective is to approximate the unknown conditional distribution $p^\star(\mathbf{Y}_{\mathrm{tar}}\mid\mathbf{y}_{\mathrm{ctx}})$.

An external probabilistic forecaster with fixed learned parameters $\phi$ estimates the conditional quantiles of each $Y_h$ given $\mathbf{y}_{\mathrm{ctx}}$. For quantile levels $\mathcal{Q}=\{\tau_1,\ldots,\tau_Q\}$ with $0<\tau_1<\cdots<\tau_Q<1$, we collect these estimates in
\begin{equation}
\mathbf{Q}_{\phi}(\mathbf{y}_{\mathrm{ctx}})
=
\left[
\hat{Q}_{\phi,h}\!\left(
\tau_k\mid\mathbf{y}_{\mathrm{ctx}}
\right)
\right]_{\substack{h=1,\ldots,H\\k=1,\ldots,Q}}
\in\mathbb{R}^{H\times Q},
\label{eq:quantile-control-object}
\end{equation}
where $\hat{Q}_{\phi,h}(\tau_k\mid\mathbf{y}_{\mathrm{ctx}})$ is the predicted $\tau_k$-quantile for future step $h$. These estimates may be predicted directly or computed from forecast samples; any sampling randomness is suppressed in the notation.

We use $\mathbf{Q}_{\phi}(\mathbf{y}_{\mathrm{ctx}})$ as a \emph{distributional control}: an explicit set of marginal predictive summaries across the forecast horizon. For example, with levels $\{0.1,0.5,0.9\}$, the median describes predicted location, the gap between the outer quantiles describes spread, and their relative distances from the median can indicate asymmetry at these levels. These summaries do not specify dependence across future time steps.

We study how these controls can be incorporated into diffusion forecasting. For mechanisms that condition the denoising network on the quantile matrix, we write the predictive distribution as
\begin{equation}
p_{\theta}\!\left(
\mathbf{Y}_{\mathrm{tar}}
\mid
\mathbf{y}_{\mathrm{ctx}},
\mathbf{Q}_{\phi}(\mathbf{y}_{\mathrm{ctx}});\,
\Psi
\right),
\label{eq:distributional-control}
\end{equation}
where $\Psi$ denotes the integration mechanism and $\theta$ collects the parameters of the resulting forecasting model for that mechanism. Output-level ensembling mixes the two predictive distributions directly and does not take $\mathbf{Q}_{\phi}(\mathbf{y}_{\mathrm{ctx}})$ as a network input.

\subsection{Quantile Integration Mechanisms}
\label{subsec:interaction-levels}

We examine output-level ensembling and three mechanisms for incorporating quantile controls within the denoising network, illustrated in Figure~\ref{fig:distributional_control}.

\textbf{Output-level ensembling.} We form a linear pool of the diffusion and external predictive distributions, with the mixing weight selected by validation CRPS. Both forecasters remain unchanged, and their forecasts are combined at the output without modifying the denoising process.

\textbf{Input-level conditioning.} We concatenate the quantile trajectories with the noisy denoiser input at each diffusion step. Quantile information enters through the input layer and propagates through the backbone's subsequent representations. The reported variant fine-tunes the backbone so its weights can adapt to the added quantile channels.

\textbf{Feature-level modulation.} Feature-wise linear modulation (FiLM) \citep{FiLMPerez} maps the quantile controls to scale and shift parameters that modulate intermediate denoiser features. We train the modulation parameters and fine-tune the backbone to adapt the denoiser to the altered intermediate representations.

\textbf{Residual control.} At each diffusion step, a trainable temporal controller processes the quantile trajectories conditioned on the timestep. Its block outputs pass through zero-initialized $1 \times 1$ projections and are added to the inputs of the corresponding residual blocks in the frozen backbone. We compare injection before the first block, the first two blocks, or all residual blocks, yielding one-block, two-block, and full-depth variants, respectively.

Section~\ref{subsec:interaction-study} compares these mechanisms. The following subsection describes Control-TS, our implementation of full-depth residual control.

\subsection{Control-TS: Residual Quantile Control}
\label{subsec:architecture}
\label{subsec:frozen-backbone}

We instantiate full-depth residual control as Control-TS on TSDiff \citep{kollovieh2023tsdiff}, an unconditional diffusion model whose architecture builds on SSSD \citep{alcaraz2022sssd}, which adapts DiffWave \citep{kong2020diffwave} using S4 temporal blocks \citep{gu2022efficiently}. For Control-TS, we fix $\Psi$ to full-depth residual control and $\theta=(\theta_B,\theta_C)$, where $\theta_B$ denotes the pretrained diffusion-backbone parameters and $\theta_C$ denotes the control-pathway parameters, including its input projection and zero-initialized residual projections. TSDiff learns a noise predictor $\epsilon_{\theta_B}(x_t,t)$, where $x_t$ denotes the noisy context-future sequence at diffusion step $t$. Forecasting conditions reverse diffusion on the observed context through observation self-guidance \citep{kollovieh2023tsdiff}. We freeze $\theta_B$ and optimize only $\theta_C$. Full DDPM and guidance equations are given in Appendix~\ref{app:diffusion}.

\begin{figure}[t]
    \centering
    \includegraphics[width=0.84\textwidth]{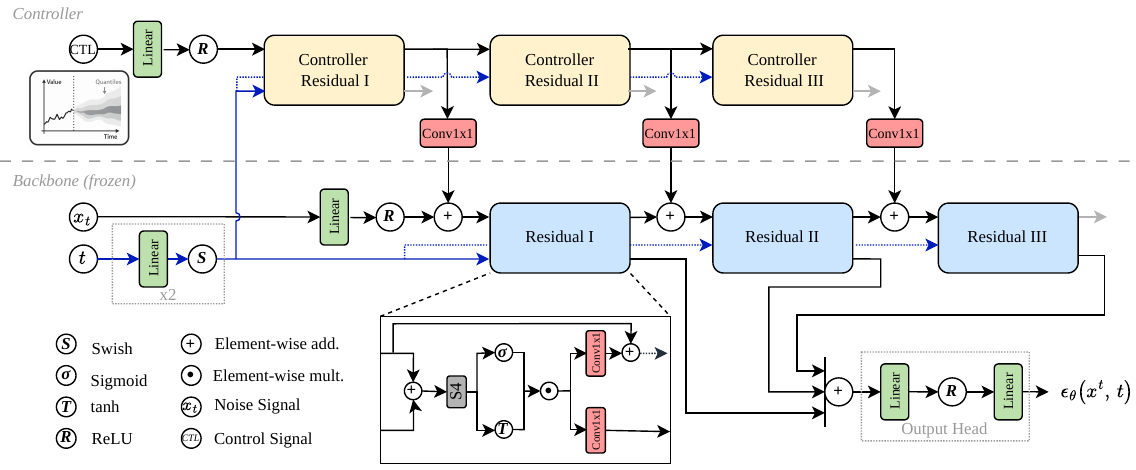}
    \caption{\textbf{Control-TS, the full-depth residual realization.} A dedicated quantile pathway injects depth-aligned control through zero-initialized $1 \times 1$ projections into the frozen backbone.} %
    \label{fig:control_ts_architecture}
\end{figure}

Control-TS separates generation from control: a dedicated pathway learns how predicted quantiles should influence the frozen backbone at every residual block (Figure~\ref{fig:control_ts_architecture}).

The controller comprises three randomly initialized residual blocks matching the backbone's residual-stack architecture (Appendix~\ref{app:architecture-details}). Let $B_{\theta_B}^{(k)}$ and $C_{\theta_C}^{(k)}$ denote the corresponding backbone and Controller blocks, $Z_{\theta_C}^{(k)}$ zero-initialized $1\times1$ projections \citep{nichol2021improvedddpm,controlnet}, and $\tau(t)$ the timestep embedding. The input projections $E_{\theta_B}$ and $E_{\theta_C}$ map the noisy sequence $x_t$ and control signal $c$ to their respective hidden representations. With initial states $h_t^{(0)}=E_{\theta_B}(x_t)$ and $u_t^{(0)}=E_{\theta_C}(c)$, the block updates for $k=1,2,3$ are
\begin{equation}
\begin{aligned}
u_t^{(k)} &= C_{\theta_C}^{(k)}\!\left(u_t^{(k-1)},\tau(t)\right),\\
h_t^{(k)} &= B_{\theta_B}^{(k)}\!\left(h_t^{(k-1)}+Z_{\theta_C}^{(k)}(u_t^{(k)}),\tau(t)\right).
\end{aligned}
\label{eq:controlts-recurrence}
\end{equation}
Because every $Z_{\theta_C}^{(k)}$ initially outputs zero, the controlled noise predictor exactly matches the pretrained backbone's noise predictor at initialization. Section~\ref{subsec:conditioning_ablation} evaluates the contributions of zero initialization and full-depth control.

\subsection{Training and Forecast Generation}
\label{subsec:control-signal}
\label{subsec:training-sampling}

We instantiate the external forecaster with DeepAR \citep{salinas2020deepar} or TFT \citep{lim2021tft} in GluonTS \citep{alexandrov2020gluonts}. Q1 uses $\mathcal{Q}=\{0.5\}$, Q3 uses $\{0.1,0.5,0.9\}$, and Q5 uses $\{0.1,0.25,0.5,0.75,0.9\}$. To align control with the full diffusion sequence, we replicate the scaled observed context across the $Q$ channels and append the predicted horizon quantiles, forming $c\in\mathbb{R}^{(L_{\mathrm{ctx}}+H)\times Q}$. When lag features are enabled, the same lag expansion is applied to the diffusion input and control. Appendices~\ref{app:control-models} and \ref{app:preproc} give extraction, scaling, alignment, and lag details.

During training, only $\theta_C$ is optimized under the standard diffusion noise-prediction objective; the backbone and external forecaster remain fixed. At inference, independent Gaussian draws are propagated through controlled reverse diffusion, with the original backbone-only observation self-guidance enforcing context consistency. Algorithms~\ref{alg:controlts-train} and \ref{alg:controlts-sample} (Appendix~\ref{app:algorithms}) provide complete pseudocode; Appendices~\ref{app:denoiser-loss} and \ref{app:val-crps-curves} show training curves.

\section{Experiments}
\label{sec:experiments}
We first compare quantile integration mechanisms and examine the design of residual control. We then evaluate Control-TS against established probabilistic forecasters, assess its gains under comparable parameter budgets, and test its applicability to CSDI and MG-TSD. Next, we investigate the role of the control information by comparing predicted quantiles with observed-context controls and varying quantile granularity and predictor choice. We subsequently assess robustness to degraded controls, the downstream forecasting utility of generated futures, and computational cost.

Unless otherwise stated, \emph{Control-TS} denotes full-depth residual control with the external forecaster (DeepAR or TFT) and quantile set (Q1, Q3, or Q5) selected by validation CRPS on each dataset. Complete configuration-level results are provided in Appendix~\ref{app:variant_results}.
Appendix ~\ref{app:experiments} provides full experimental details, with the protocols and hyperparameters specified in Appendix \ref{app:hparams}.

\textbf{Datasets and evaluation.} Our main experiments evaluate \emph{univariate probabilistic forecasting} on eight real-world benchmarks from the GluonTS repository \citep{alexandrov2020gluonts}: Electricity \citep{dua2017uci}, Exchange \citep{lai2018lstnet}, KDDCup \citep{kddcup2018,zhang2020kddcupdata}, M4-Hourly \citep{MAKRIDAKIS202054}, Solar \citep{lai2018lstnet}, Traffic \citep{dua2017uci}, UberTLC \citep{fivethirtyeight2016uber}, and Wikipedia \citep{gasthaus2019sqf}. Dataset statistics, splits, and licenses are documented in Appendices~\ref{app:datasets} and \ref{app:asset-licenses}. We evaluate probabilistic accuracy using the continuous ranked probability score (CRPS) \citep{gneiting2007scoring,hersbach2000crps}, computed from 100 sampled forecast paths; lower values indicate better performance. Appendix~\ref{app:crps} provides the scoring and sampling details. Unless otherwise stated, we report the mean and standard deviation over three independent runs with different random seeds.

\subsection{Effect of the integration mechanism}
\label{subsec:interaction-study}
\label{subsec:conditioning_ablation}

We compare the integration mechanisms in Section~\ref{subsec:interaction-levels}, holding the external forecaster fixed within each dataset and using identical quantile levels across quantile-conditioned variants. Concatenation and FiLM fine-tune the backbone, whereas residual variants keep it frozen. The ensembling weight is selected by validation CRPS. We vary residual depth by injecting corrections before the first block, the first two blocks, or all three blocks (Control-TS). We report fine-tuned variants of concatenation and FiLM, allowing the denoiser to adapt to the introduced quantile conditioning. We also evaluated frozen-backbone variants of both mechanisms, but observed worse performance. Full results are in Appendix~\ref{app:frozen-conditioning}.
TSDiff's training-free refinement \citep{kollovieh2023tsdiff} operates on forecast samples rather than quantile trajectories, so we evaluate it with DeepAR in Section~\ref{subsec:forecasting-performance}.

\begin{table}[ht]
\centering
\caption{\textbf{Quantile integration mechanisms.} Best values are \textbf{bold}, second-best values are \underline{underlined}.}
\label{tab:interaction_levels}
\scriptsize
\setlength{\tabcolsep}{2.5pt}
\resizebox{\textwidth}{!}{%
\begin{tabular}{lccccccc}
\toprule
Dataset & Backbone & \shortstack{Output\\ensembling}
& \shortstack{Input\\concat.} & FiLM
& \shortstack{1-block\\residual}
& \shortstack{2-block\\residual}
& \shortstack{Control-TS\\(full-depth)} \\
\midrule
Electricity & \underline{0.049$\pm$0.000} & \underline{0.049$\pm$0.000} & 0.062$\pm$0.000 & 0.054$\pm$0.000 & 0.052$\pm$0.000 & 0.052$\pm$0.000 & \textbf{0.047$\pm$0.000} \\
Exchange & 0.012$\pm$0.001 & \underline{0.008$\pm$0.000} & \textbf{0.007$\pm$0.000} & \underline{0.008$\pm$0.000} & 0.009$\pm$0.000 & \underline{0.008$\pm$0.000} & \underline{0.008$\pm$0.000} \\
KDDCup & 0.312$\pm$0.001 & 0.320$\pm$0.003 & 0.289$\pm$0.002 & \underline{0.268$\pm$0.005} & 0.286$\pm$0.001 & 0.285$\pm$0.002 & \textbf{0.249$\pm$0.001} \\
M4-Hourly & 0.034$\pm$0.000 & 0.034$\pm$0.000 & 0.027$\pm$0.001 & \underline{0.026$\pm$0.001} & 0.032$\pm$0.000 & \underline{0.026$\pm$0.000} & \textbf{0.022$\pm$0.000} \\
Solar & 0.402$\pm$0.000 & 0.378$\pm$0.001 & \textbf{0.308$\pm$0.004} & 0.360$\pm$0.002 & 0.444$\pm$0.000 & 0.415$\pm$0.001 & \underline{0.341$\pm$0.000} \\
Traffic & \underline{0.099$\pm$0.000} & \underline{0.099$\pm$0.000} & 0.117$\pm$0.000 & 0.116$\pm$0.001 & 0.125$\pm$0.000 & 0.118$\pm$0.000 & \textbf{0.097$\pm$0.000} \\
UberTLC & 0.172$\pm$0.000 & \textbf{0.163$\pm$0.000} & 0.187$\pm$0.001 & 0.184$\pm$0.001 & 0.203$\pm$0.001 & 0.193$\pm$0.001 & \underline{0.169$\pm$0.000} \\
Wikipedia & 0.211$\pm$0.001 & \underline{0.208$\pm$0.000} & 0.241$\pm$0.001 & 0.235$\pm$0.001 & 0.248$\pm$0.002 & 0.245$\pm$0.001 & \textbf{0.200$\pm$0.000} \\
\bottomrule
\end{tabular}%
}
\end{table}

Table~\ref{tab:interaction_levels} shows that ensembling, concatenation, and FiLM each improve backbone CRPS on four datasets, while the one- and two-block residual variants each improve three. Full-depth residual control is the only mechanism to improve all eight, and it is best or tied-best on five. Concatenation attains the lowest CRPS on Exchange and Solar, and ensembling on UberTLC. Within the residual variants, full-depth control outperforms both shallower designs, except for a tie with the two-block variant on Exchange after rounding. Appendix~\ref{app:interaction-summary} provides the complete summary.

Zero-initialized residual projections preserve the pretrained backbone's predictions at the start of training. Appendix~\ref{app:arch-ablations} provides further ablations of projection initialization, controller initialization, and controller depth.

\subsection{Forecasting performance and applicability}
\label{subsec:forecasting-performance}
\label{subsec:backbone-transfer}

\textbf{Forecasting accuracy.} Table~\ref{tab:main_results} compares Control-TS with established forecasters across eight GluonTS benchmarks using CRPS under a shared univariate protocol. Details of our K$^2$VAE and CN-Diff evaluations are provided in Appendix~\ref{app:additional-baselines}. Control-TS selects its external forecaster by validation CRPS for each dataset and quantile set, and it improves on TSDiff across all eight benchmarks, with the largest reduction of approximately 35\% on M4-Hourly (0.034 to 0.022). It achieves the best or tied-best mean CRPS among all evaluated forecasters on five benchmarks; TSFlow performs better on Electricity, Traffic, and UberTLC. Control-TS also outperforms the external forecaster on most datasets, supporting the value of integrating its predictions with the diffusion backbone.

\setlength{\textfloatsep}{12pt}
\begin{table}[t]
\centering
\caption{\textbf{Forecasting results on eight benchmarks.} CRPS; lower is better. Best/second-best values are \textbf{bold}/\underline{underlined}, including ties. Our evaluations report mean$\pm$SD over three independent runs with different random seeds.}
\label{tab:main_results}
\scriptsize
\setlength{\tabcolsep}{1.5pt}
\begin{tabular}{@{}l!{\color{black!18}\vrule}c!{\color{black!18}\vrule}c!{\color{black!18}\vrule}c!{\color{black!18}\vrule}c!{\color{black!18}\vrule}c!{\color{black!18}\vrule}c!{\color{black!18}\vrule}c!{\color{black!18}\vrule}c@{}}
\toprule
Method & Electricity & Exchange & KDDCup & M4-Hourly & Solar & Traffic & UberTLC & Wikipedia \\
\midrule
ARIMA & 0.344$\pm$0.000 & \textbf{0.008$\pm$0.000} & 0.514$\pm$0.000 & 0.031$\pm$0.000 & 0.558$\pm$0.003 & 0.486$\pm$0.000 & 0.478$\pm$0.000 & 0.654$\pm$0.000 \\
DLinear & 0.058$\pm$0.001 & 0.015$\pm$0.004 & 0.318$\pm$0.015 & 0.055$\pm$0.007 & 0.794$\pm$0.027 & 0.131$\pm$0.000 & 0.250$\pm$0.006 & 0.259$\pm$0.002 \\
DeepAR    & 0.051$\pm$0.000 & 0.013$\pm$0.004 & 0.362$\pm$0.017 & 0.046$\pm$0.006 & 0.429$\pm$0.055 & 0.103$\pm$0.002 & 0.169$\pm$0.002 & 0.215$\pm$0.003 \\

TFT       & 0.060$\pm$0.001 & 0.011$\pm$0.000 & 0.543$\pm$0.048 & 0.038$\pm$0.002 & 0.371$\pm$0.006 & 0.128$\pm$0.005 & 0.202$\pm$0.009 & 0.219$\pm$0.004 \\

PatchTST  & 0.055$\pm$0.001 & \underline{0.010$\pm$0.001} & 0.420$\pm$0.011 & 0.034$\pm$0.004 & 0.728$\pm$0.015 & 0.151$\pm$0.007 & 0.219$\pm$0.004 & 0.209$\pm$0.001 \\
iTransformer & 0.074$\pm$0.000 & \underline{0.010$\pm$0.000} & - & - & 0.496$\pm$0.000 & 0.158$\pm$0.000 & - & 0.262$\pm$0.000 \\

\midrule

K$^2$VAE & 0.078$\pm$0.000 & 0.014$\pm$0.000 & 0.332$\pm$0.000 & 0.046$\pm$0.000 & 0.371$\pm$0.000 & 0.126$\pm$0.000 & 0.178$\pm$0.001 & 0.291$\pm$0.000 \\

CSDI      & 0.051$\pm$0.000 & 0.013$\pm$0.001 & 0.309$\pm$0.006 & 0.043$\pm$0.004 & 0.360$\pm$0.006 & 0.152$\pm$0.001 & 0.213$\pm$0.007 & 0.318$\pm$0.012 \\

SSSD      & 0.048$\pm$0.001 & \underline{0.010$\pm$0.001} & 0.274$\pm$0.009 & 0.050$\pm$0.007 & 0.384$\pm$0.023 & 0.097$\pm$0.002 & \underline{0.156$\pm$0.007} & 0.209$\pm$0.004 \\

Bilo\v{s} et al. & 0.067$\pm$0.002 & 0.012$\pm$0.004 & 1.147$\pm$0.300 & - & 0.379$\pm$0.009 & 0.317$\pm$0.053 & 0.450$\pm$0.086 & 0.318$\pm$0.022 \\

CN-Diff & 0.084$\pm$0.000 & 0.011$\pm$0.000 & 0.470$\pm$0.000 & 0.050$\pm$0.000 & 0.535$\pm$0.002 & 0.173$\pm$0.000 & 0.471$\pm$0.000 & 0.397$\pm$0.000 \\

TSDiff & 0.049$\pm$0.000 & 0.012$\pm$0.001 & 0.312$\pm$0.001 & 0.034$\pm$0.000 & 0.402$\pm$0.000 & 0.099$\pm$0.000 & 0.172$\pm$0.000 & 0.211$\pm$0.001 \\

TSDiff ML-Q & 0.049$\pm$0.000 & 0.011$\pm$0.000 & 0.360$\pm$0.002 & 0.038$\pm$0.000 & 0.399$\pm$0.000 & 0.094$\pm$0.000 & 0.172$\pm$0.000 & 0.215$\pm$0.000 \\

TSDiff LMC-Q & 0.051$\pm$0.000 & 0.011$\pm$0.001 & 0.361$\pm$0.002 & 0.039$\pm$0.000 & 0.400$\pm$0.000 & 0.097$\pm$0.000 & 0.172$\pm$0.001 & 0.215$\pm$0.000 \\

\midrule

TSFlow-Ucond. & 0.049$\pm$0.000 & 0.011$\pm$0.001 & 0.287$\pm$0.005 & 0.032$\pm$0.001 & 0.396$\pm$0.005 & \underline{0.086$\pm$0.000} & \textbf{0.154$\pm$0.002} & 0.291$\pm$0.005 \\

TSFlow-Cond. (OU) & \textbf{0.045$\pm$0.000} & \textbf{0.008$\pm$0.000} & 0.288$\pm$0.004 & 0.028$\pm$0.008 & \underline{0.344$\pm$0.006} & \textbf{0.082$\pm$0.000} & \textbf{0.154$\pm$0.002} & 0.207$\pm$0.001 \\
\midrule

DeepAR-Large & \underline{0.047$\pm$0.001} & 0.022$\pm$0.006 & 0.380$\pm$0.019 & 0.047$\pm$0.006 & 0.371$\pm$0.020 & 0.092$\pm$0.005 & 0.164$\pm$0.005 & 0.227$\pm$0.006 \\
TFT-Large & 0.048$\pm$0.000 & \underline{0.010$\pm$0.001} & 0.354$\pm$0.044 & 0.030$\pm$0.003 & 0.552$\pm$0.045 & 0.099$\pm$0.010 & 0.161$\pm$0.002 & 0.205$\pm$0.001 \\
TSDiff-Wide & 0.048$\pm$0.000 & 0.012$\pm$0.000 & 0.345$\pm$0.002 & 0.035$\pm$0.000 & 0.409$\pm$0.001 & 0.095$\pm$0.000 & 0.167$\pm$0.000 & 0.222$\pm$0.000 \\
\midrule

\textbf{Control-TS-Q1}
& 0.049$\pm$0.000
& \textbf{0.008$\pm$0.000}
& 0.282$\pm$0.001
& \underline{0.025$\pm$0.000}
& 0.408$\pm$0.000
& 0.098$\pm$0.000
& 0.170$\pm$0.000
& 0.203$\pm$0.000 \\

\textbf{Control-TS-Q3}
& 0.048$\pm$0.000
& \textbf{0.008$\pm$0.000}
& \textbf{0.249$\pm$0.001}
& 0.026$\pm$0.000
& 0.392$\pm$0.000
& 0.097$\pm$0.000
& 0.169$\pm$0.000
& \underline{0.202$\pm$0.000} \\

\textbf{Control-TS-Q5}
& \underline{0.047$\pm$0.000}
& \textbf{0.008$\pm$0.000}
& \underline{0.256$\pm$0.001}
& \textbf{0.022$\pm$0.000}
& \textbf{0.341$\pm$0.000}
& 0.100$\pm$0.000
& 0.172$\pm$0.000
& \textbf{0.200$\pm$0.000} \\
\bottomrule
\end{tabular}%
\end{table}

\textbf{Training-free refinement.} TSDiff's refinement mode combines an external forecaster with the frozen backbone without training: it refines the forecaster's samples under the backbone's learned density \citep{kollovieh2023tsdiff}. Initialized with DeepAR samples, its two quantile-regularized variants (TSDiff ML-Q and LMC-Q; Appendix~\ref{app:tsdiff-refine}) improve on TSDiff only on Exchange, Solar and Traffic, whereas Control-TS outperforms both on every dataset except Traffic. This suggests that learning how quantiles condition denoising offers benefits beyond post-hoc refinement.

\textbf{Capacity controls.} TSDiff-Wide doubles the backbone width from 64 to 128, giving 498-502k parameters. Complete Control-TS uses 408-506k, including its external forecaster; its diffusion component alone uses 380-385k. Control-TS achieves lower CRPS than TSDiff-Wide on six datasets, while TSDiff-Wide performs better on Traffic and UberTLC. It also matches or outperforms both DeepAR-Large and TFT-Large on six datasets. These comparisons support gains beyond simply increasing parameter count (Appendix~\ref{app:capacity-matched}).

\begin{figure}[ht]
    \centering
    \includegraphics[width=0.96\textwidth]{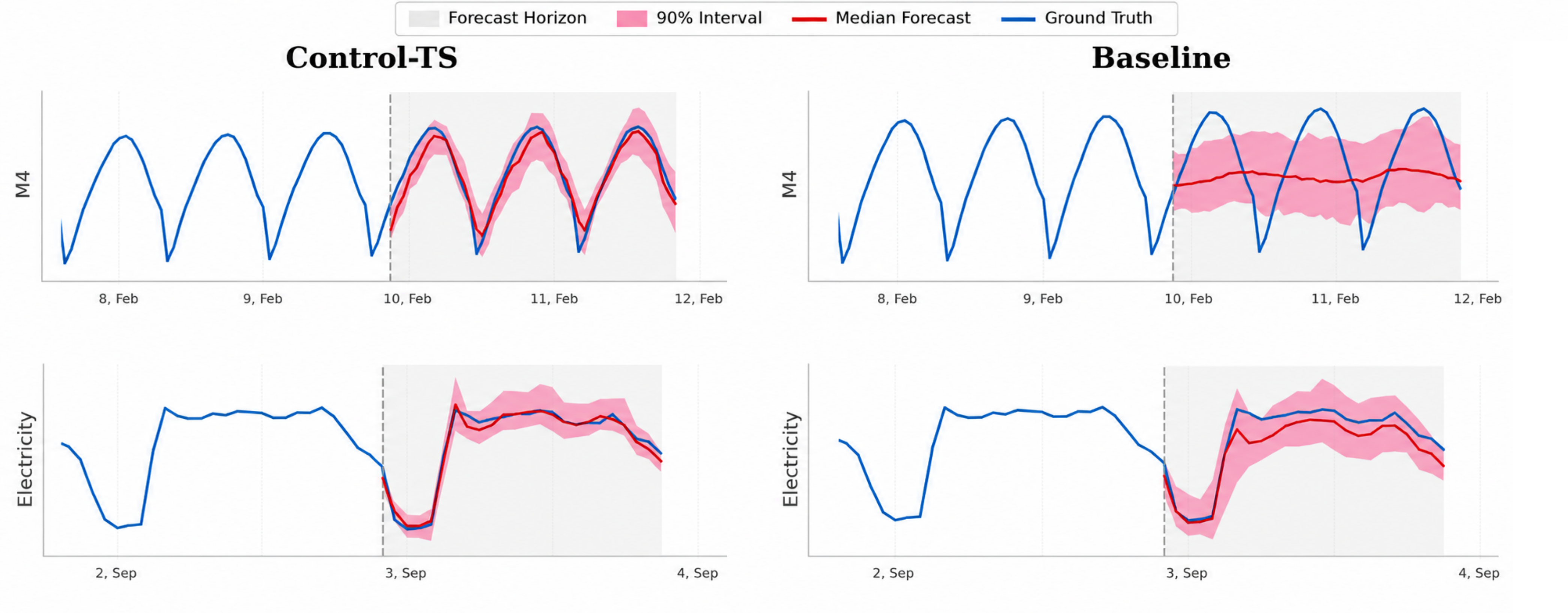}
    \caption{\textbf{Qualitative predictive distributions.} Control-TS and TSDiff medians and 90\% intervals on shared M4-Hourly and Electricity contexts. Shaded regions denote the forecast horizon.}
    \label{fig:qualitative_uncertainty_comparison}
    \label{fig:forecast_comparison}
\end{figure}

\textbf{Qualitative forecasts.} In Figure~\ref{fig:qualitative_uncertainty_comparison}, Control-TS continues the oscillatory pattern in the M4-Hourly example, where TSDiff's median flattens, and yields a sharper, better-centered interval in the Electricity example. Appendix~\ref{app:qualitative-examples} provides further comparisons.

\textbf{Applicability to additional backbones.} We assess applicability beyond TSDiff using CSDI \citep{tashiro2021csdi} with DeepAR controls and multivariate MG-TSD \citep{fan2024mgtsd} with DeepVAR controls \citep{salinas2019deepvar}. Each backbone retains its native forecasting computation while incorporating the quantile-driven residual pathway. Quantile control reduces CSDI's CRPS by 0.34\%-11.05\% across all eight datasets and improves MG-TSD's CRPS-sum on five of eight (Table~\ref{tab:backbone-transfer}). These results extend the observed benefits beyond TSDiff, with their magnitude and consistency depending on the backbone and dataset.

\begin{table}[ht]
\centering
\caption{\textbf{Quantile control on additional diffusion backbones.} CRPS for univariate CSDI; CRPS-sum for multivariate MG-TSD. Results are based on three independent runs with different random seeds. Positive $\Delta$ denotes relative improvement over the corresponding backbone. Scores are not pooled across backbones or with the main benchmark comparison. Appendix~\ref{app:backbone-transfer} provides detailed results and protocols.}
\label{tab:backbone-transfer}
\small
\setlength{\tabcolsep}{4pt}
\begin{tabular}{lccc}
\toprule
Backbone & Metric & Improved datasets & Range $\Delta$ (\%) \\
\midrule
CSDI & CRPS & 8/8 & $[+0.34,+11.05]$ \\
MG-TSD & CRPS-sum & 5/8 & $[-7.23,+38.27]$ \\
\bottomrule
\end{tabular}
\end{table}

\subsection{Role of the control information}
\label{subsec:quantile-richness}

Having examined how controls enter the denoising network, we next investigate what information they provide. We compare predicted quantiles with observed-context controls, then examine quantile granularity and the choice of external forecaster.

\textbf{Observed context as control.} We keep the same full-depth residual architecture and replace the predicted quantiles with the observed context repeated to the same number of channels: once for Q1, three times for Q3, and five times for Q5. This comparison is intended to isolate the quantile forecasts, so that a difference is not attributable to the architecture or to access to the history alone. Across widths, predicted quantiles give lower CRPS in 21 of the 24 dataset--granularity comparisons. The repeated context is lower for Electricity Q1, Solar Q1, and Traffic Q5 (Appendix~\ref{app:context-control}).

\textbf{Quantile granularity.} We vary the granularity of the explicit forecast through Q1, Q3, and Q5 (Table~\ref{tab:main_results}). Validation selects Q3 or Q5 on every dataset, with large gains on KDDCup, M4-Hourly, and Solar. Multi-quantile variants do not uniformly improve over Q1, however, and increasing $Q$ changes both information content and input width. The results therefore support richer quantile representations without isolating the contributions of location, dispersion, and asymmetry.

\textbf{External forecaster.} The complete DeepAR/TFT $\times$ Q1/Q3/Q5 grid in Appendix~\ref{app:variant_results} shows that the preferred control source depends on the dataset. TFT supplies the selected controls on five benchmarks, while DeepAR supplies them on M4-Hourly, Solar, and Traffic.

\subsection{Robustness to degraded controls}
\label{subsec:control_signal_quality}

\begin{figure}[ht]
    \centering
    \begin{subfigure}[t]{0.39\textwidth}
        \centering
        \includegraphics[height=0.147\textheight]{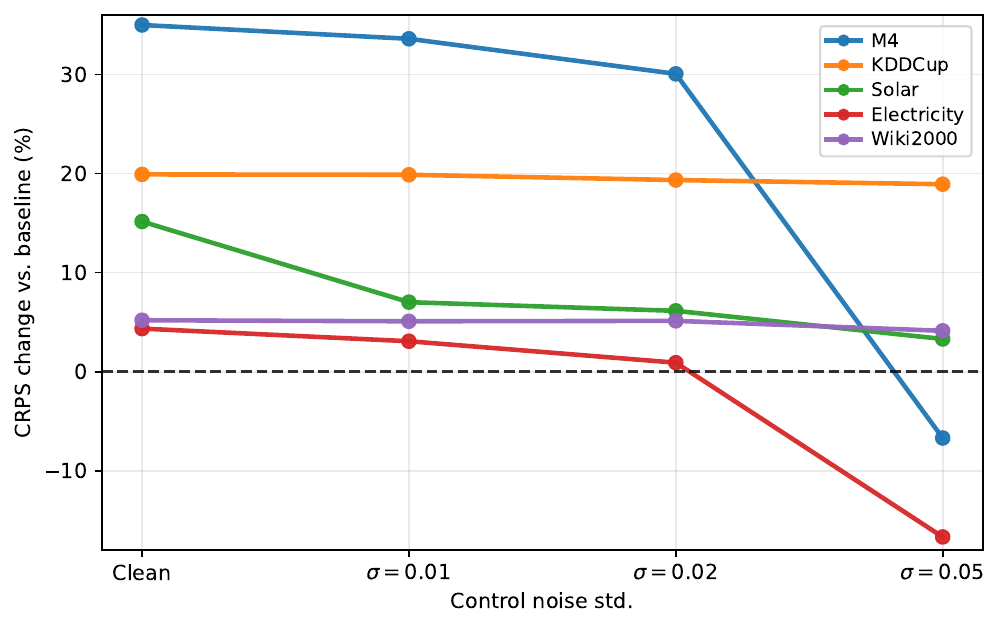}
        \caption{Gaussian control perturbations.}
        \label{fig:gaussian_noise_robustness}
    \end{subfigure}\hfill
    \begin{subfigure}[t]{0.60\textwidth}
        \centering
        \includegraphics[height=0.147\textheight]{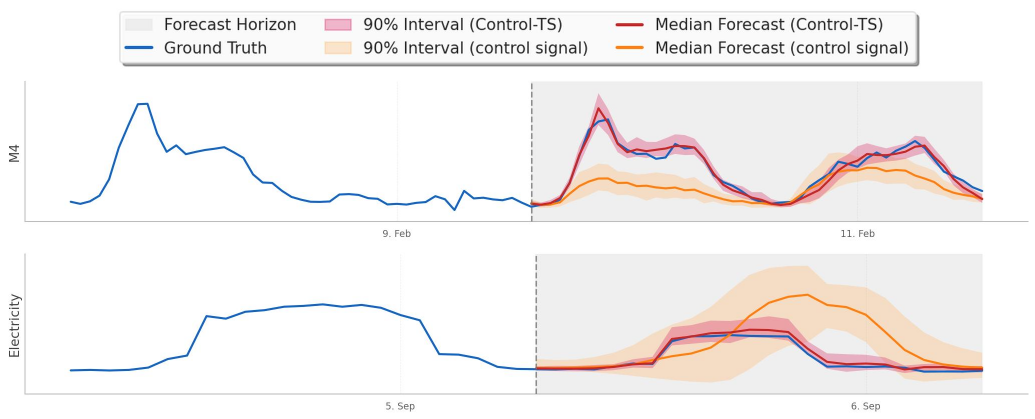}
        \caption{Examples under misaligned control signals.}
        \label{fig:buffered_control}
    \end{subfigure}
    \caption{\textbf{Robustness to degraded control signals.} Left: CRPS improvement over TSDiff as Gaussian corruption increases; positive values indicate retained gains. Right: selected M4-Hourly and Electricity examples with inaccurate quantile scaffolds; Control-TS stays closer to the realized continuation than the scaffold. Full corruption results and failure cases are in Appendices~\ref{app:robustness}, \ref{app:bad_control_failure}.}
    \label{fig:natural_control_behaviors}
\end{figure}

\textbf{Gaussian corruption.} We assess sensitivity to control quality by adding Gaussian noise to the control signal at inference in normalized control space, with the controller and backbone fixed. Figure~\ref{fig:gaussian_noise_robustness} reports the percentage CRPS improvement over the backbone on representative datasets; positive values indicate retained gains. Performance degrades smoothly: gains are preserved under mild to moderate corruption but become less consistent under stronger noise, as expected when the quantile signal becomes weakly informative. Appendix~\ref{app:robustness} provides the full results.

\textbf{Naturally misaligned controls.} We also examine cases in which the predicted quantile scaffold is poorly aligned with the realized future. The examples in Figure~\ref{fig:buffered_control} show that Control-TS does not simply track this unreliable scaffold: its samples remain shaped by the diffusion backbone while using the quantile estimates as guidance. Together, the Gaussian-corruption results and the examples with naturally low-quality controls indicate that Control-TS uses the auxiliary uncertainty as guidance rather than as a hard target. Appendix~\ref{app:bad_control_failure} includes failure cases in which the forecast follows a misaligned scaffold.

\subsection{Forecasting signal in generated futures}
\label{subsec:predictive-utility}

Under Train-on-Hybrid, Test-on-Real (THTR) \citep{articletstr}, a forecaster trains on real contexts paired with generated futures and is evaluated on held-out real data using CRPS. CLPS uses a shared ridge evaluator with inexpensive, deterministic fitting for fixed data to assess the forecasting signal accessible to that evaluator. The \emph{Linear} block of Table~\ref{tab:clps_thtr} reports CLPS; DeepAR and TFT probe nonlinear forecasters using the same training pairs. Appendix~\ref{app:clps-details} provides the protocol.

\setlength{\textfloatsep}{20pt plus 2pt minus 4pt}
\begin{table}[t]
\centering
\caption{\textbf{Train-on-Hybrid, Test-on-Real CRPS.} Mean$\pm$SD over three independent runs with different random seeds. The better synthetic source is \textbf{bold}, including ties.}
\label{tab:clps_thtr}
\label{tab:clps_main}
\scriptsize
\setlength{\tabcolsep}{1.5pt}
\begin{tabular}{@{}clcccccccc@{}}
\toprule
& & Electricity & Exchange & KDDCup & M4-Hourly & Solar & Traffic & UberTLC & Wikipedia \\
\midrule
\multirow{2}{*}{\shortstack{Linear\\(CLPS)}} & TSDiff
& \textbf{0.070$\pm$0.001}
& 0.096$\pm$0.000
& 0.900$\pm$0.001
& 0.041$\pm$0.000
& 0.543$\pm$0.002
& 0.157$\pm$0.001
& \textbf{0.303$\pm$0.003}
& 0.392$\pm$0.006 \\
& Control-TS
& 0.072$\pm$0.004
& \textbf{0.028$\pm$0.000}
& \textbf{0.699$\pm$0.006}
& \textbf{0.038$\pm$0.001}
& \textbf{0.526$\pm$0.003}
& \textbf{0.154$\pm$0.001}
& 0.306$\pm$0.001
& \textbf{0.374$\pm$0.007} \\
\midrule
\multirow{2}{*}{DeepAR} & TSDiff
& \textbf{0.054$\pm$0.002}
& \textbf{0.011$\pm$0.003}
& 0.480$\pm$0.027
& \textbf{0.025$\pm$0.001}
& \textbf{0.380$\pm$0.016}
& 0.116$\pm$0.002
& 0.174$\pm$0.002
& 0.224$\pm$0.006 \\
& Control-TS
& \textbf{0.054$\pm$0.001}
& 0.012$\pm$0.008
& \textbf{0.411$\pm$0.033}
& 0.035$\pm$0.007
& 0.387$\pm$0.009
& \textbf{0.110$\pm$0.002}
& \textbf{0.173$\pm$0.006}
& \textbf{0.223$\pm$0.005} \\
\midrule
\multirow{2}{*}{TFT} & TSDiff
& \textbf{0.058$\pm$0.001}
& 0.015$\pm$0.002
& 0.818$\pm$0.099
& \textbf{0.032$\pm$0.002}
& 0.374$\pm$0.007
& 0.137$\pm$0.007
& 0.238$\pm$0.011
& 0.232$\pm$0.003 \\
& Control-TS
& 0.060$\pm$0.001
& \textbf{0.007$\pm$0.000}
& \textbf{0.632$\pm$0.001}
& \textbf{0.032$\pm$0.001}
& \textbf{0.364$\pm$0.007}
& \textbf{0.124$\pm$0.014}
& \textbf{0.221$\pm$0.010}
& \textbf{0.230$\pm$0.001} \\
\bottomrule
\end{tabular}%
\end{table}

Control-TS improves CLPS over TSDiff on six datasets, most strongly on Exchange and KDDCup, with small losses on Electricity and UberTLC. The DeepAR and TFT evaluations show gains on four and six datasets, respectively, each with one additional tie.

\subsection{Computational cost}
\label{subsec:compute-overhead}

\textbf{Forecast generation.} We compare full 100-step forecast calls at batch size 16, including online quantile generation. Control-TS increases wall-clock time by 17-61\% (30\% on average), FLOPs by 7-25\% (16.5\% on average), and peak VRAM by 2-7\%. The interface therefore trades additional computation for improved forecasting, with comparatively modest inference-memory overhead.

\textbf{Controller training.} During controller fitting with cached controls, per-step time overhead is below 15\% on six datasets and below 66\% on all eight; peak VRAM increases by approximately 65\%. These training costs are incremental to the pretrained backbone and external forecaster. Appendix~\ref{app:compute} provides the full accounting.

\section{Related work}
\textbf{Explicit probabilistic forecasting.} Classical and neural forecasters expose uncertainty through likelihoods, quantiles, or intervals \citep{MAKRIDAKIS202054,smyl2020hybrid}. Examples include ARIMA \citep{hyndman2008forecast}, DeepAR \citep{salinas2020deepar}, TFT \citep{lim2021tft}, DLinear \citep{zeng2023dlinear}, PatchTST \citep{nie2023patchtst}, iTransformer \citep{liu2024itransformer}, and global state-space or factorized models \citep{rangapuram2018deepstate,wang2019deepfactors,sen2019deepglo}. We study predicted quantiles as controls for a separate implicit forecaster, rather than only as final summaries.

\textbf{Implicit generative forecasting.} Diffusion modeling \citep{sohl2015nonequilibrium,ho2020ddpm,song2021score_sde,song2021ddim} supports autoregressive forecasting \citep{rasul2021arddm}, score-based imputation \citep{tashiro2021csdi,alcaraz2022sssd}, and guided trajectory generation \citep{bilos2023,kollovieh2023tsdiff}; TSFlow uses flow matching \citep{kollovieh2025flow}. Recent models include K$^2$VAE \citep{wu2025k2vae} and CN-Diff \citep{rishi2025cndiff}. Others condition diffusion on point forecasts: a pretrained mean estimator in CARD \citep{han2022card}, transformer forecasts in TMDM \citep{li2024tmdm}, and linear autoregressive initialization in TimeDiff \citep{shen2023timediff}. TSDiff refines base forecasts under its learned density without training \citep{kollovieh2023tsdiff}. We instead learn to condition a frozen backbone on externally predicted quantile trajectories and compare integration mechanisms, including TSDiff refinement (Section~\ref{subsec:forecasting-performance}).

\textbf{Conditioning and control.} Classifier and classifier-free guidance steer sampling \citep{dhariwal2021diffusionbeatgans,ho2022classifierfreediffusionguidance}; FiLM modulates features \citep{FiLMPerez}. Frozen-backbone adaptation includes parallel residual adapters \citep{rebuffi2018efficient} and side-tuning \citep{zhang2020sidetuning}. Earlier diffusion implementations also use zero-initialized output layers within residual branches \citep{nichol2021improvedddpm}. For spatial conditioning in image generation, ControlNet connects a frozen denoiser to a trainable copy of its pretrained encoder and middle block through zero-initialized projections \citep{controlnet}. Control-TS builds on this auxiliary residual-conditioning principle, while initializing its controller independently of the pretrained backbone. The temporal controller maps horizon-aligned quantile forecasts and the diffusion timestep to residual corrections throughout the frozen denoising network.  Our study focuses on predicted quantiles as controls that convey marginal forecast uncertainty, investigating how this information can effectively guide diffusion forecasting through a comparison of alternative integration mechanisms.

\textbf{Evaluation and downstream forecasting.} We use CRPS for forecasting evaluation \citep{gneiting2007scoring,hersbach2000crps}. CLPS adapts TSDiff's linear predictive score \citep{kollovieh2023tsdiff} to real-context/synthetic-future pairs in Train-on-Hybrid, Test-on-Real evaluation \citep{articletstr}, testing the forecasting signal in generated futures.

\section{Conclusion}
\label{sec:conclusion}
We established predicted quantile trajectories as explicit distributional controls for implicit diffusion-based trajectory generation, comparing input conditioning, feature modulation, residual control, and output-level ensembling. Full-depth residual control (Control-TS) yielded the most consistent improvements on the primary backbone, motivating ablations and analyses of quantile granularity, predictor choice, controller design, robustness, and computational cost. Control-TS is competitive with strong probabilistic forecasters, reducing backbone CRPS on all eight benchmarks by up to approximately 35\% and outperforming a matched-capacity diffusion baseline on six. Experiments with CSDI and MG-TSD support broader applicability, with benefits varying across backbones and datasets. We introduce CLPS, a Train-on-Hybrid, Test-on-Real (THTR) criterion, to complement forecast accuracy by assessing the downstream forecasting utility of generated futures.

\section{Limitations and Future Work}
\label{sec:limitations}
Our study focuses on quantile trajectories predicted by learned forecasters. Inaccurate or misaligned controls may still degrade forecasting performance. The use of other predictive summaries (e.g., moments or temporal correlations), domain-expert knowledge, and system constraints as control signals remains unexplored. Future work will investigate how to represent and incorporate these signals, alongside adaptive gating to adjust the strength of quantile guidance at each injection.

\subsection*{AI Use Statement}

Generative AI tools were used to assist with drafting and polishing the manuscript, including improving grammar, clarity, concision, and organization, and to assist with the refinement and organization of existing code. The authors reviewed and revised the AI-assisted text and code to ensure that the manuscript accurately reflected the methods, experimental findings, and conclusions. The authors take full responsibility for the final manuscript, including the accuracy of its claims, results, and references.

\bibliography{Control-TS}
\bibliographystyle{preprint}

\clearpage

\appendix

\section{Appendices}
\label{sec:appendix}
\subsection{Diffusion Equations, Schedules, and Observation Self-Guidance}
\label{app:diffusion}
We provide the complete DDPM formulation \citep{ho2020ddpm}, including the forward and reverse diffusion processes, schedule definitions, the training objective, and the explicit formulation of \textbf{observation self-guidance} used during inference. Observation self-guidance was introduced in TSDiff \citep{kollovieh2023tsdiff}; we restate it here for completeness and to fix notation in the controlled setting.

\textbf{Forward process and schedule.} TSDiff follows a standard forward diffusion process
\begin{align}
q(x_t \mid x_0) &= \mathcal{N}\!\big(\sqrt{\bar{\alpha}_t}\,x_0,\; (1-\bar{\alpha}_t)I\big), \\
x_t &= \sqrt{\bar{\alpha}_t}\,x_0 + \sqrt{1-\bar{\alpha}_t}\,\epsilon,\qquad \epsilon \sim \mathcal{N}(0,I),
\label{eq:forward}
\end{align}
where the schedule is defined by $\beta_t \in (0,1)$, $\alpha_t = 1-\beta_t$, and $\bar{\alpha}_t=\prod_{s=1}^{t}\alpha_s$.

\textbf{DDPM training objective.} The unconditional backbone is trained using the standard DDPM objective
\begin{align}
\mathcal{L}_{\mathrm{diff}}(\theta_B) &= \mathbb{E}_{x_0,\,t,\,\epsilon}\Big[\big\|\epsilon - \epsilon_{\theta_B}(x_t,t)\big\|_2^2\Big], \label{eq:ddpm-loss} \\
&\text{where } t \sim \mathrm{Unif}\{1,\dots,T_{\mathrm{diff}}\}, \ \epsilon\sim\mathcal{N}(0,I). \nonumber
\end{align}

\textbf{Reverse process.} Reverse diffusion follows the standard DDPM update. Let $\epsilon_\theta(x_t,t)$ be the noise predictor and define the denoised estimate
\begin{equation}
\tilde{x}_0(x_t,t) \;=\; \frac{x_t - \sqrt{1-\bar{\alpha}_t}\,\epsilon_\theta(x_t,t)}{\sqrt{\bar{\alpha}_t}}.
\label{eq:x0hat}
\end{equation}
We form the Gaussian reverse kernel mean
\begin{equation}
\mu_\theta(x_t,t) \;=\; \frac{1}{\sqrt{\alpha_t}}\Big(x_t - \frac{\beta_t}{\sqrt{1-\bar{\alpha}_t}}\,\epsilon_\theta(x_t,t)\Big),
\label{eq:ddpm-mean}
\end{equation}
and sample
\begin{equation}
x_{t-1} \;=\; \mu_\theta(x_t,t) + \sqrt{\tilde{\beta}_t}\,z,\qquad z\sim\mathcal{N}(0,I),
\label{eq:ddpm-step}
\end{equation}
where $\tilde{\beta}_t = \beta_t \frac{1-\bar{\alpha}_{t-1}}{1-\bar{\alpha}_t}$ is the standard posterior variance. For the backbone, these expressions use $\epsilon_{\theta_B}(x_t,t)$; for Control-TS, they use the controlled predictor $\hat{\epsilon}_{\theta}(x_t,t\mid c)$ with $\theta=(\theta_B,\theta_C)$.

\textbf{Observation self-guidance.} Observation self-guidance enforces consistency with the observed context by modifying the reverse-kernel mean using a guidance score \citep{kollovieh2023tsdiff}. Let $m\in\{0,1\}^{L\times D}$ be an observation mask with $m_{i,d}=1$ for context time indices $i\in\{1,\dots,L_{\mathrm{ctx}}\}$ and $m_{i,d}=0$ for forecast time indices $i\in\{L_{\mathrm{ctx}}+1,\dots,L_{\mathrm{ctx}}+L_{\mathrm{tar}}\}$ (when lag features are enabled, $D>1$ includes the corresponding lag channels). In Control-TS, observation self-guidance is evaluated through the unmodified frozen backbone, with the Controller disabled. Thus, the denoised estimate used in the guidance energy is the backbone-only estimate $\tilde{x}_{0,B}(x_t,t)$. Given an observation tensor $y$ (context + zero-padded future), the energy is
\begin{equation}
E(x_t,t;y,m) \;=\;
\sum_{(i,d):\,m_{i,d}=1}
\ell\!\Big(\tilde{x}_{0,B}(x_t,t)_{i,d},\, y_{i,d}\Big).
\label{eq:self_guidance_energy}
\end{equation}
where $\ell$ is either mean-squared error (MSE guidance) or the quantile loss used by our sampler (quantile guidance) \citep{koenker1978quantiles}. The corresponding guidance score is the negative energy gradient,
\begin{equation}
g(x_t,t) \;=\; -\nabla_{x_t} E(x_t,t;y,m),
\label{eq:obs-score}
\end{equation}
which induces a guided reverse kernel of the form
\begin{equation}
p_{\theta_B,\theta_C}(x_{t-1}\mid x_t,y,m,c)
\;=\;
\mathcal{N}\!\Big(
\mu_{\theta_B,\theta_C}(x_t,t,c)
+
\lambda_t\, g(x_t,t),
\;
\tilde{\beta}_t I
\Big),
\label{eq:obs-guidance-kernel}
\end{equation}
with $\lambda_t = s\,\tilde{\beta}_t$ for a user-chosen guidance scale $s$, matching observation self-guidance in \citet{kollovieh2023tsdiff}. Here, $\mu_{\theta_B,\theta_C}$ is computed using the controlled denoising prediction, whereas $g(x_t,t)$ is computed from the backbone-only denoised estimate $\tilde{x}_{0,B}(x_t,t)$ in Eq.~\ref{eq:self_guidance_energy}.

\subsection{Extended Control-TS Formalization and Architectural Details}
\label{app:architecture-details}

This subsection expands the main-text formalization of Control-TS and provides additional architectural details referenced in Section~\ref{subsec:architecture}.

\textbf{Backbone and Controller streams.} Let $\{B_{\theta_B}^{(k)}\}_{k=1}^{K}$ denote the residual blocks of the frozen diffusion backbone, $\{C_{\theta_C}^{(k)}\}_{k=1}^{K}$ the corresponding Controller residual blocks, and $\{Z_{\theta_C}^{(k)}\}_{k=1}^{K}$ the zero-initialized channel projection operators associated with each residual depth. We define the backbone hidden stream $\{h_t^{(k)}\}_{k=0}^{K}$ and the Controller stream $\{u_t^{(k)}\}_{k=0}^{K}$ at diffusion step $t$, where $h_t^{(k)}$ (resp.\ $u_t^{(k)}$) denotes the representation after the $k$-th backbone (resp.\ Controller) residual block.

\textbf{Controller initialization.} Controller residual blocks are copied during model construction and randomly initialized.

\textbf{Time embeddings and input projections.} We use a shared timestep embedding $\tau_{\theta_B}(t)$ that is injected into every residual block in both branches. Separately, $E_{\theta_B}(\cdot)$ denotes the backbone input projection applied to the currently noised trajectory $x_t$, and $E_{\theta_C}(\cdot)$ denotes the Controller input projection applied to the control signal $c$. The two streams are initialized as
\begin{equation}
h_t^{(0)} = E_{\theta_B}(x_t),
\qquad
u_t^{(0)} = E_{\theta_C}(c).
\label{eq:app-init}
\end{equation}

\textbf{Depth-aligned residual injection.} For $k=1,\dots,K$, the Controller and backbone evolve as
\begin{align}
u_t^{(k)} &= C_{\theta_C}^{(k)}\!\big(u_t^{(k-1)},\, \tau_{\theta_B}(t)\big), \label{eq:app-ctrl-rec}\\
h_t^{(k)} &= B_{\theta_B}^{(k)}\!\Big(h_t^{(k-1)} + Z_{\theta_C}^{(k)}\!\big(u_t^{(k)}\big),\, \tau_{\theta_B}(t)\Big). \label{eq:app-backbone-rec}
\end{align}
The control signal enters only through $u_t^{(0)}=E_{\theta_C}(c)$, after which the Controller produces a depth-aligned sequence of residual states $\{u_t^{(k)}\}$ whose projected residuals are injected additively into the corresponding backbone block $B_{\theta_B}^{(k)}$.

\textbf{Noise prediction head.} Each backbone residual block emits a local latent contribution $s_t^{(k)}$, which is aggregated and mapped to the final noise prediction via the backbone readout head (the final convolutional head in Fig.~\ref{fig:control_ts_architecture}). Writing $s_t=\sum_{k=1}^{K}s_t^{(k)}$, the controlled predictor is
\begin{equation}
\hat{\epsilon}_{\theta}(x_t,t \mid c)=\mathrm{OutputHead}_{\theta_B}(s_t).
\label{eq:app-output}
\end{equation}

\textbf{Exact reduction to the frozen backbone when control is ``off''.} Define the backbone-only predictor as the controlled model evaluated with all zero-initialized channel projections set to zero:
\begin{equation}
\epsilon_{\theta_B}(x_t,t)
\;:=\;
\hat{\epsilon}_{\theta}(x_t,t \mid c)\Big|_{Z_{\theta_C}^{(1)}=\cdots=Z_{\theta_C}^{(K)}\equiv 0}.
\label{eq:app-backbone}
\end{equation}
Because each $Z_{\theta_C}^{(k)}$ is initialized to output zero, Control-TS satisfies
\begin{equation}
\hat{\epsilon}_{\theta}(x_t,t\mid c)=\epsilon_{\theta_B}(x_t,t)
\qquad \text{at initialization,}
\label{eq:app-init-reduction}
\end{equation}
i.e., it reduces exactly to the pretrained diffusion backbone before learning any control influence. A zero-initialized pointwise output projection likewise appears in the official implementation of \citet{nichol2021improvedddpm} (\texttt{proj\_out} in \url{https://github.com/openai/improved-diffusion/blob/main/improved_diffusion/unet.py}).

\textbf{Interpretation of the control mechanism.} For interpretation, let $\delta_t^{(k)} := Z_{\theta_C}^{(k)}(u_t^{(k)})$ denote the control perturbation injected at depth $k$. Since the backbone parameters are frozen, the Controller learns to inject these $\delta_t^{(k)}$ vectors to steer the backbone hidden-state trajectory. Although the Controller and backbone are coupled and evaluated in a single forward pass at each diffusion step, it is useful to interpret the resulting aggregated latent representation $s_t$ as the sum of a baseline component and a control-induced deviation:
\begin{equation}
\hat{\epsilon}_{\theta}(x_t,t\mid c)=
\mathrm{OutputHead}_{\theta_B}\Bigg(
\underbrace{s_t(\emptyset)}_{\text{Baseline state}}
\;+\;
\underbrace{\sum_{k=1}^{K}\Delta s_t^{(k)}(\delta_t^{(1:k)})}_{\text{Steering effect}}
\Bigg).
\label{eq:app-additive}
\end{equation}
Here, $s_t(\emptyset)$ is the aggregated latent representation the pretrained backbone would produce without control, and $\Delta s_t^{(k)}$ captures the generally non-linear deviation at depth $k$ induced by the cumulative injections. Equation~\eqref{eq:app-additive} is intended as an interpretation of the control mechanism, not as an alternative computational graph.

\subsection{External Forecasters and Quantile Generation}
\label{app:control-models}
The control signal is produced by an external probabilistic forecaster implemented in GluonTS. We consider two families: \textbf{DeepAR} and the \textbf{Temporal Fusion Transformer (TFT)}. Both are trained per dataset to produce probabilistic forecasts over the target horizon of length $L_{\mathrm{tar}}$ from the observed context.

\textbf{Quantiles and variants.} At inference time, we extract a set of quantile trajectories over the horizon of length $L_{\mathrm{tar}}$ and use them as channels in the control signal. The quantile set is configurable; in our experiments we use Q1 $\{0.5\}$, Q3 $\{0.1,0.5,0.9\}$, or Q5 $\{0.1,0.25,0.5,0.75,0.9\}$ depending on the dataset and ablation setting.

\textbf{Quantile extraction.} Given a trained external forecaster, we draw a moderate number of forecast samples and compute the desired quantiles from the resulting predictive distribution. Concretely, for each requested level $q$ we obtain the trajectory in $\mathbb{R}^{L_{\mathrm{tar}}}$ by evaluating the predictor's quantile function $f\mapsto f.\mathrm{quantile}(q)$.

\subsection{Preprocessing, Alignment, and Lag Indices}
\label{app:preproc}
\textbf{Scaling.} We use per-series mean-absolute scaling with a small floor for numerical stability. In the diffusion model, scaling is applied internally to the context before concatenating context and horizon for diffusion; the same scale is used to normalize both the noising target and any auxiliary quantities. For external forecasters, we similarly normalize each series by its mean absolute value computed from the observed context portion before fitting the probabilistic forecaster.

\textbf{Control alignment.} Let $Q$ be the number of control quantiles. The external forecaster outputs horizon quantiles in scaled space in $\mathbb{R}^{L_{\mathrm{tar}}\times Q}$; we form the aligned control signal $c \in \mathbb{R}^{L\times Q}$ with $L=L_{\mathrm{ctx}}+L_{\mathrm{tar}}$ by replicating the scaled context across $Q$ channels and appending the horizon quantiles.

\textbf{Lag augmentation.} When lag features are enabled for some datasets, we use the default GluonTS lag values for the corresponding data frequency \citep{alexandrov2020gluonts} to incorporate historical information beyond the modeled window of length $L=L_{\mathrm{ctx}}+L_{\mathrm{tar}}$. We expand both the diffusion input and the control signal with the same lag indices to preserve temporal alignment; the resulting lag features contribute $C\!-\!1$ additional channels. For hourly data, where \(C\) is the total number of channels after lag augmentation, we use lags $\{24,48,72,96,120,144,168,336,504,672\}$; for daily/business-day data, we use $\{30,60,90,120,150,180,210,240,270,300,330,360\}$. For multi-quantile control, lag expansion is applied per quantile channel and then concatenated across quantiles.

\newpage

\subsection{Denoiser Loss Reduction Under Control}
\label{app:denoiser-loss}
\begin{figure}[H]
\centering
\includegraphics[width=0.9\linewidth,height=0.28\textheight,keepaspectratio]{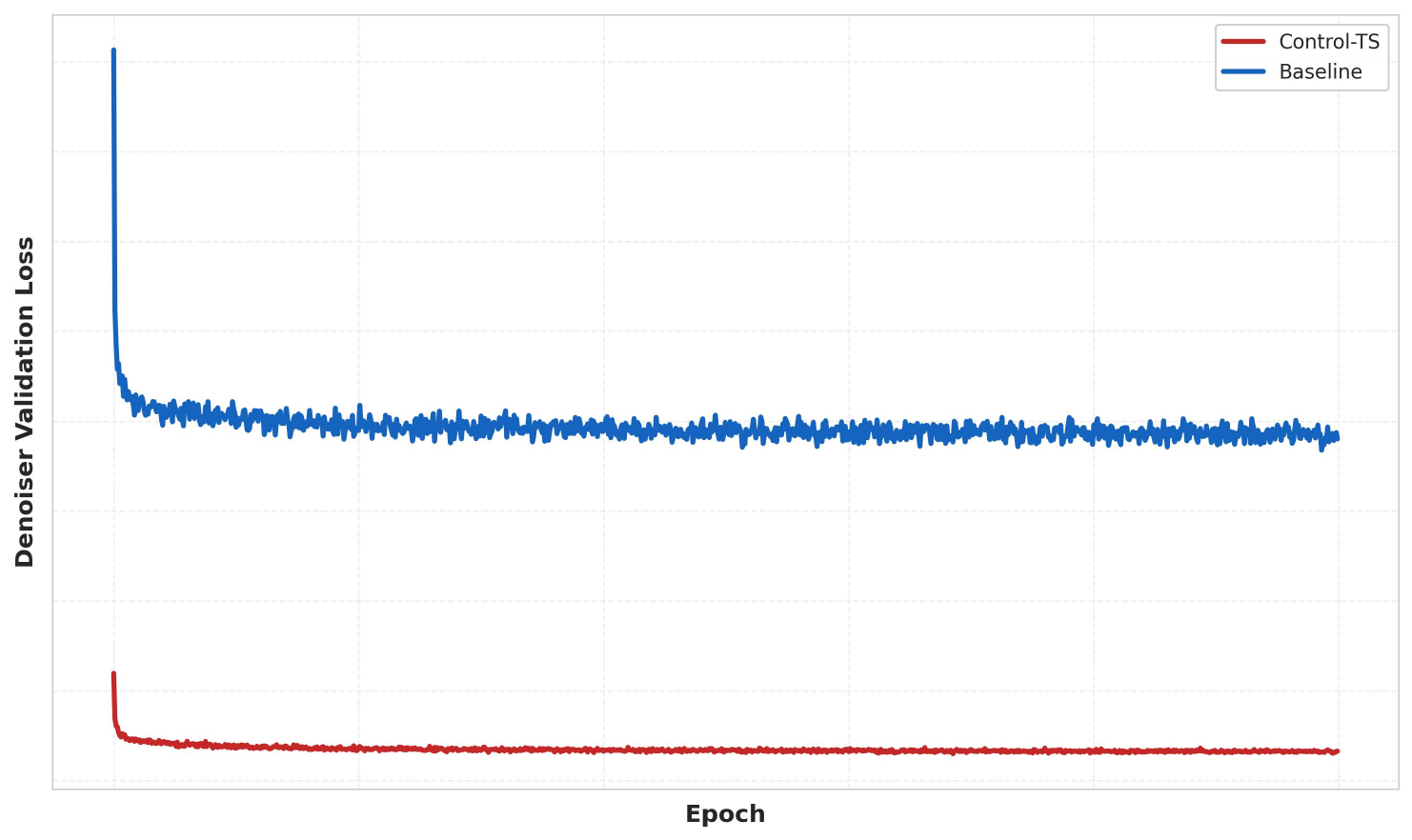}
\caption{Wikipedia denoiser L2 loss over 1000 epochs.}
\label{fig:denoiser-loss}
\end{figure}

We plot the denoiser loss on Wikipedia over 1000 training epochs in Fig.~\ref{fig:denoiser-loss}. Training starts from the exact pretrained TSDiff backbone (control injections are zero at initialization) and only improves as control is introduced; the injected uncertainty-aware scaffold makes the remaining denoising residual easier to learn, often yielding a rapid early drop (visible after the first few epochs). We observe a similar pattern across the other datasets.

\subsection{Validation CRPS Curves Across Training}
\label{app:val-crps-curves}
\begin{figure}[H]
\centering
\includegraphics[width=0.49\linewidth]{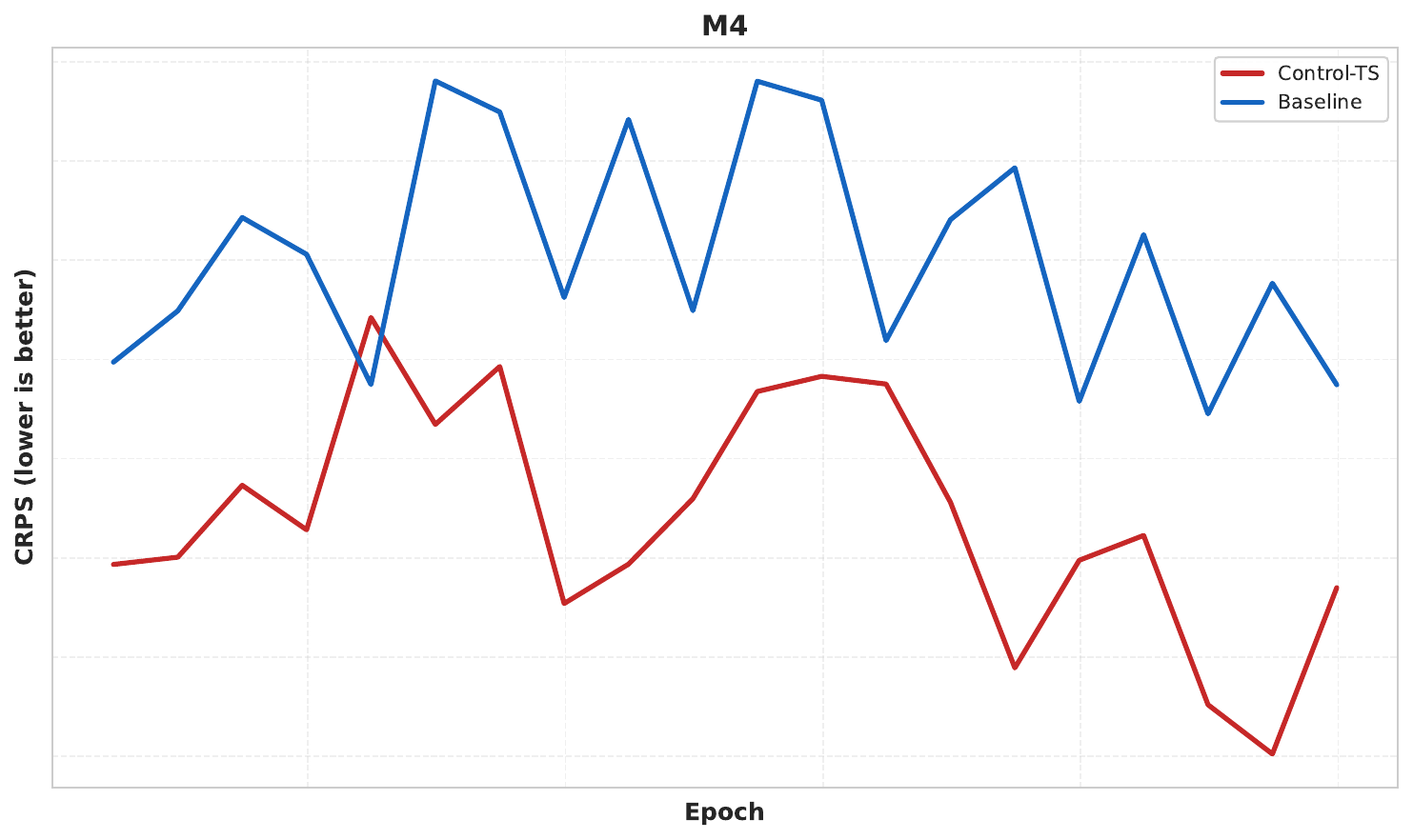}\includegraphics[width=0.49\linewidth]{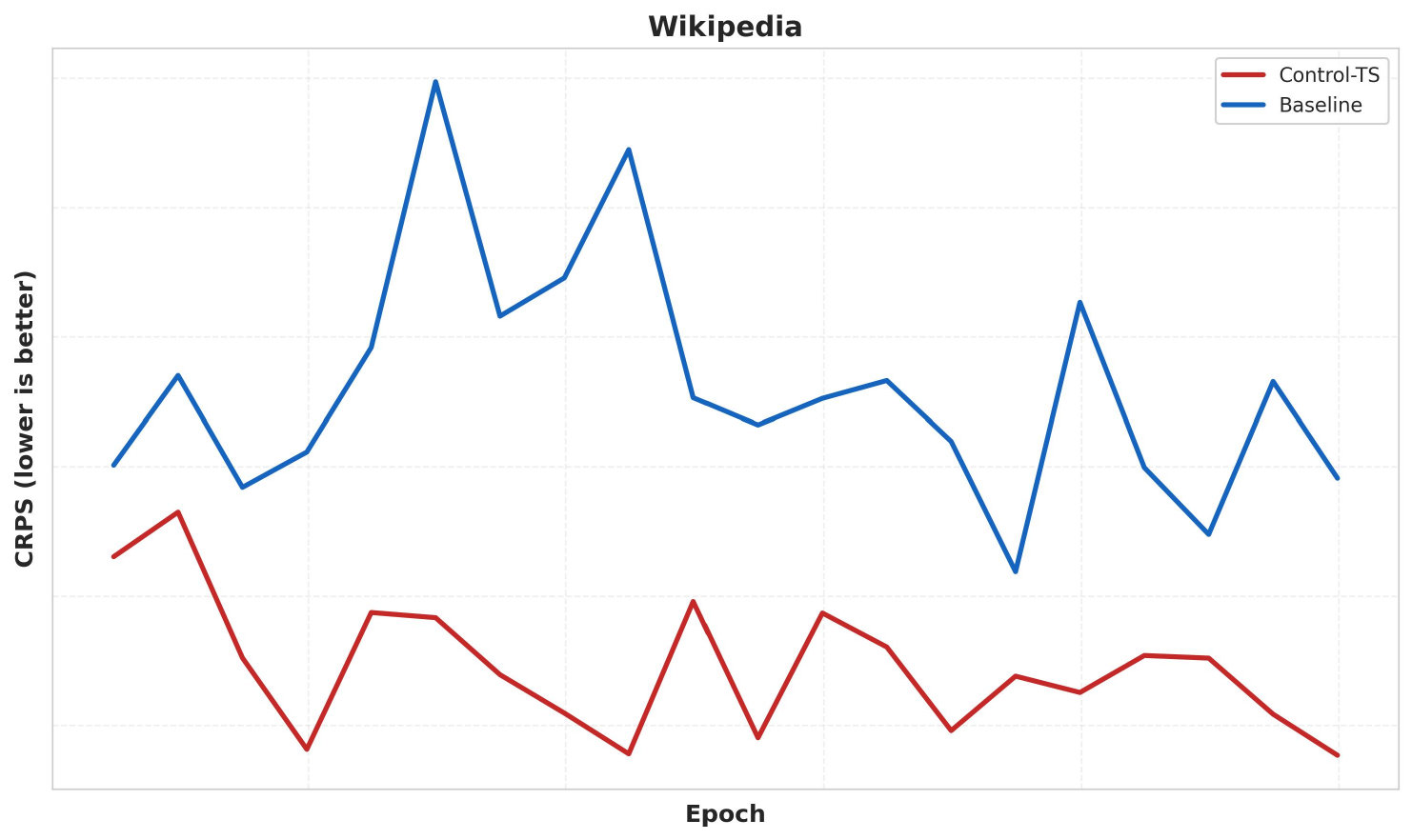}
\caption{Validation CRPS evaluated every 50 epochs during 1000-epoch Controller training (two example datasets).}
\label{fig:val-crps-curves}
\end{figure}

We plot validation CRPS during the 1000-epoch Controller training for two representative datasets in Fig.~\ref{fig:val-crps-curves}, evaluated every 50 epochs. Unlike full test-set evaluation (Table~\ref{tab:hparams}), these periodic validation evaluations use fewer Monte Carlo forecast samples for efficiency. We observe a similar pattern across the other datasets.

\newpage

\subsection{Algorithms (Training and Sampling Pseudocode)}
\label{app:algorithms}
We provide pseudocode for Control-TS training and inference. The training procedure is summarized in \hyperref[alg:controlts-train]{Algorithm~\ref*{alg:controlts-train}}, and inference via controlled reverse diffusion is summarized in \hyperref[alg:controlts-sample]{Algorithm~\ref*{alg:controlts-sample}}.

\begin{algorithm}[H]
\caption{Control-TS training.}
\label{alg:controlts-train}
\begin{algorithmic}[1]
\STATE \textbf{Input:} training set; $T_{\mathrm{diff}}$; quantile set $\mathcal{Q}$; pretrained diffusion backbone $\theta_B$; pretrained external-forecaster.
\STATE Freeze $\theta_B$.
\STATE Initialize controller blocks using the construction-time initialization described in Appendix~\ref{app:architecture-details}; initialize all injection-projection weights and biases to zero.
\FOR{each optimization step}
\STATE Sample batch $(x_0, y_{1:L_{\mathrm{ctx}}})$.
\STATE Sample $t\sim\mathrm{Unif}\{1,\dots,T_{\mathrm{diff}}\}$ and $\epsilon\sim\mathcal{N}(0,I)$.
\STATE $x_t \leftarrow \sqrt{\bar{\alpha}_t}\,x_0 + \sqrt{1-\bar{\alpha}_t}\,\epsilon$.
\STATE $c \leftarrow \mathrm{Align}(\mathrm{Quantiles}(\text{external-forecaster}(y_{1:L_{\mathrm{ctx}}}),\mathcal{Q})) \in \mathbb{R}^{L\times Q}$, $L=L_{\mathrm{ctx}}+L_{\mathrm{tar}}$.
\STATE $\hat{\epsilon} \leftarrow \epsilon_{\theta_B,\theta_C}(x_t,t\mid c)$.
\STATE $\mathcal{L}_{\mathrm{denoise}} \leftarrow \|\epsilon-\hat{\epsilon}\|_2^2$.
\STATE Update $\theta_C$ by one Adam step on $\nabla_{\theta_C}\mathcal{L}_{\mathrm{denoise}}$, holding $\theta_B$ fixed.
\ENDFOR
\end{algorithmic}
\end{algorithm}

\begin{algorithm}[H]
\caption{Control-TS inference.}
\label{alg:controlts-sample}
\begin{algorithmic}[1]
\STATE \textbf{Input:} context $y_{1:L_{\mathrm{ctx}}}$; quantile set $\mathcal{Q}$; pretrained external-forecaster; the frozen diffusion backbone $\theta_B$ and learned $\theta_C$; mask $m$; $N$.
\STATE $c \leftarrow \mathrm{Align}(\mathrm{Quantiles}(\text{external-forecaster}(y_{1:L_{\mathrm{ctx}}}),\mathcal{Q}))$.
\STATE $y \leftarrow [y_{1:L_{\mathrm{ctx}}};\,0_{1:L_{\mathrm{tar}}}]$ (context + zero-padded future; Appendix~\ref{app:diffusion}).
\STATE \textbf{for} $n=1,2,\dots,N$ \textbf{do}
\STATE $x_{T_{\mathrm{diff}}}^{(n)} \sim \mathcal{N}(0,I)$ on $\mathbb{R}^{L\times D}$ with $L=L_{\mathrm{ctx}}+L_{\mathrm{tar}}$.
\FOR{$t=T_{\mathrm{diff}},T_{\mathrm{diff}}-1,\dots,1$}
\STATE $\hat{\epsilon}\leftarrow \hat{\epsilon}(x_t^{(n)},t\mid c)$.
\STATE $\mu_\theta \leftarrow \mu_\theta(x_t^{(n)},t;\hat{\epsilon})$ (Eq.~\ref{eq:ddpm-mean}).
\STATE $g \leftarrow g(x_t^{(n)},t;y,m)$ (Eq.~\ref{eq:obs-score}).
\STATE $\tilde{\mu}_\theta \leftarrow \mu_\theta + \lambda_t g$ (Eq.~\ref{eq:obs-guidance-kernel}, $\lambda_t=s\,\tilde{\beta}_t$).
\STATE $z\sim\mathcal{N}(0,I)$ if $t>1$ else $z\leftarrow 0$.
\STATE $x_{t-1}^{(n)} \leftarrow \tilde{\mu}_\theta + \sqrt{\tilde{\beta}_t}\,z$.
\ENDFOR
\STATE $\hat{\mathbf{y}}_{\mathrm{tar}}^{(n)} \leftarrow$ target-horizon segment of $x_0^{(n)}$.
\STATE \textbf{end for}
\STATE \textbf{Return:} $\{\hat{\mathbf{y}}_{\mathrm{tar}}^{(n)}\}_{n=1}^{N}$.
\end{algorithmic}
\end{algorithm}

\noindent\textit{Pretrained modules.} The diffusion backbone $\theta_B$ (frozen throughout Controller training) and the external forecaster are inputs to both procedures; in practice they are taken from off-the-shelf checkpoints in GluonTS \citep{alexandrov2020gluonts} or other libraries, or trained beforehand and loaded from disk. Only $\theta_C$ is optimized in \hyperref[alg:controlts-train]{Algorithm~\ref*{alg:controlts-train}}.

\noindent\textit{Notation.} $\hat{\epsilon}(x_t,t\mid c)$ is the controlled denoiser (frozen backbone $\theta_B$ plus learned controller $\theta_C$) at diffusion step $t$ with control $c$.

\newpage

\section{Experimental Details}
\label{app:experiments}

\subsection{Dataset Overview}
\label{app:datasets}
We use eight commonly used datasets and train/test splits from GluonTS \citep{alexandrov2020gluonts} spanning multiple domains and sampling frequencies. Table~\ref{tab:dataset_overview} summarizes the dataset statistics.
\begin{table}[H]
\caption{Overview of the datasets and statistics used in our experiments.}
\vspace{0.55\baselineskip}
\label{tab:dataset_overview}
\small
{\centering
\begin{tabular}{lccccccc}
\toprule
Dataset & Train & Test & Domain & Freq. & Median Seq. Len. & Pred. Len. \\
\midrule
Electricity\textsuperscript{a} & 370 & 2590 & $\mathbb{R}^+$ & H & 5833 & 24 \\
Exchange\textsuperscript{b} & 8 & 40 & $\mathbb{R}^+$ & D & 6071 & 30 \\
KDDCup\textsuperscript{c} & 270 & 270 & N & H & 10850 & 48 \\
M4-Hourly\textsuperscript{d} & 414 & 414 & N & H & 960 & 48 \\
Solar\textsuperscript{e} & 137 & 959 & $\mathbb{R}^+$ & H & 7009 & 24 \\
Traffic\textsuperscript{f} & 963 & 6741 & $(0,1)$ & H & 4001 & 24 \\
UberTLC\textsuperscript{g} & 262 & 262 & N & H & 4320 & 24 \\
Wikipedia\textsuperscript{h} & 2000 & 10000 & N & D & 792 & 30 \\
\bottomrule
\end{tabular}
\par}

\par\smallskip
{\footnotesize\raggedright
\textsuperscript{a}\url{https://archive.ics.uci.edu/ml/datasets/ElectricityLoadDiagrams20112014}\\
\textsuperscript{b}\url{https://github.com/laiguokun/multivariate-time-series-data}\\
\textsuperscript{c}\url{https://zenodo.org/record/4656756}\\
\textsuperscript{d}\url{https://github.com/Mcompetitions/M4-methods/tree/master/Dataset}\\
\textsuperscript{e}\url{https://www.nrel.gov/grid/solar-power-data.html}\\
\textsuperscript{f}\url{https://zenodo.org/record/4656132}\\
\textsuperscript{g}\url{https://github.com/fivethirtyeight/uber-tlc-foil-response}\\
\textsuperscript{h}\url{https://github.com/mbohlkeschneider/gluon-ts/tree/mv_release/datasets}
}
\end{table}

\subsection{Existing Dataset Assets and Licenses}
\label{app:asset-licenses}

We used GluonTS only to access preprocessed versions of public benchmark datasets. GluonTS is released under the Apache License 2.0 and documents that its real external datasets are included only when they have a blessed license. Table~\ref{tab:dataset_licenses} lists the dataset sources, licenses or available terms of use, and our usage. We do not redistribute raw datasets or third-party dataset archives; all datasets should be obtained from GluonTS or the original sources under their corresponding licenses and terms.

\begin{table}[H]
\centering
\caption{Existing dataset assets used in this work. Licenses and terms are reported according to the corresponding official dataset pages, repositories, or documentation pages accessed on 2026-05-07.}
\label{tab:dataset_licenses}
\scriptsize
\setlength{\tabcolsep}{2pt}
\begin{tabular}{@{}p{0.12\textwidth}p{0.27\textwidth}p{0.28\textwidth}p{0.28\textwidth}@{}}
\toprule
Dataset & GluonTS name & Source / provider & License or terms of use \\
\midrule
Solar
& \texttt{solar\_nips}
& NREL / Monash Time Series Forecasting Repository
& CC BY 4.0 via Monash/Zenodo. \\

Electricity
& \texttt{electricity\_nips}
& UCI Machine Learning Repository
& CC BY 4.0. \\

Traffic
& \texttt{traffic\_nips}
& Caltrans PeMS / Monash Time Series Forecasting Repository
& CC BY 4.0 via Monash/Zenodo. \\

Exchange
& \texttt{exchange\_rate\_nips}
& LSTNet multivariate time-series repository
& Research use stated by the source repository; no formal named license identified. \\

M4-Hourly
& \texttt{m4\_hourly}
& M4 Competition / Monash Time Series Forecasting Repository
& CC BY 4.0 via Monash/Zenodo. \\

KDDCup
& \path{kdd_cup_2018_without_missing}
& KDD Cup 2018 / Monash Time Series Forecasting Repository
& CC BY 4.0 via Monash/Zenodo. \\

UberTLC
& \texttt{uber\_tlc\_hourly}
& FiveThirtyEight / NYC Taxi \& Limousine Commission
& CC BY 4.0 unless otherwise noted by FiveThirtyEight. \\

Wikipedia
& \texttt{wiki2000\_nips}
& Wikimedia analytics datasets
& CC0 unless otherwise specified by Wikimedia. \\

GluonTS
& Dataset access utility
& Amazon / GluonTS
& Apache License 2.0. \\
\bottomrule
\end{tabular}%
\end{table}

\subsection{CRPS and Sampling Details}
\label{app:crps}
CRPS \citep{gneiting2007scoring,hersbach2000crps} is the primary proper scoring rule used for probabilistic time-series forecasting (reported as a loss; lower is better). Intuitively, CRPS measures how well the \emph{full marginal predictive distribution} at each future time step matches the realized value: it rewards forecasts that are both \emph{accurate} (well-centered) and \emph{well-calibrated} (appropriate dispersion), and it penalizes distributions that are systematically biased or mis-scaled.

For a predictive CDF $F$ and realized value $y$, CRPS is
\begin{equation}
\mathrm{CRPS}(F,y) = \int_{-\infty}^{\infty} \big(F(z) - \mathbb{I}\{z \ge y\}\big)^2\,dz.
\end{equation}
An equivalent algebraic form is
\begin{equation}
\mathrm{CRPS}(F,y) \;=\; \mathbb{E}|X-y| \;-\; \tfrac{1}{2}\mathbb{E}|X-X'|,\qquad X,X'\sim F \text{ i.i.d.}
\end{equation}
Given Monte Carlo forecast samples $\{\hat{y}^{(i)}\}_{i=1}^N$ from $F$, a standard sample-based estimator is
\begin{equation}
\widehat{\mathrm{CRPS}}(y;\{\hat{y}^{(i)}\}_{i=1}^N)
=
\frac{1}{N}\sum_{i=1}^{N}\big|\hat{y}^{(i)}-y\big|
\;-\;
\frac{1}{2N^2}\sum_{i=1}^{N}\sum_{j=1}^{N}\big|\hat{y}^{(i)}-\hat{y}^{(j)}\big|.
\label{eq:crps-sample}
\end{equation}
We use CRPS consistently for forecasting evaluation and CLPS evaluation.

For full test-set evaluation, we form the predictive distribution from $N=100$ Monte Carlo forecast samples per context; during training, periodic validation uses $N=16$ (and $N=4$ for Traffic/Wikipedia), following Table~\ref{tab:hparams}. The diffusion process uses $T=100$ steps with a linear $\beta$ schedule (from $10^{-4}$ to $10^{-1}$).

\subsection{Control-TS Hyperparameters}
\label{app:hparams}
\begin{table}[H]
\caption{Summary of key hyperparameters used by our implementation and configs.}
\vspace{0.55\baselineskip}
\label{tab:hparams}
\centering
\small
\begin{tabular}{@{}p{0.20\textwidth}p{0.73\textwidth}@{}}
\toprule
Component & Setting \\
\midrule
Hardware & NVIDIA GeForce RTX 2080 Ti \\
Diffusion steps & $T_{\mathrm{diff}}=100$ \\
$\beta$ schedule & linear from $10^{-4}$ to $10^{-1}$ \\
Backbone block & S4 residual blocks (bidirectional), $d_{\mathrm{state}}=128$ \\
Residual depth & $K=3$ (small configuration) \\
Hidden width & 64 \\
Time embedding dim & 128 \\
Optimizer & Adam \\
Learning rate & $10^{-3}$ \\
Gradient clipping & 0.5 \\
Training budget & up to 1000 epochs; periodic validation CRPS every 50 epochs \\
Batching & 128 batches per epoch; batch size 64 \\
Checkpoint selection & validation CRPS \\
Context length & 312 (KDDCup/M4-Hourly), 336 (Electricity/Solar/Traffic/UberTLC), 360 (Exchange/Wikipedia) \\
Guidance type & quantile guidance (default) \\
Guidance scale & dataset-dependent (e.g., 2) \\
Control quantiles & Q1/Q3/Q5 depending on experiment \\
Quantile extraction & 100 paths per context for DeepAR; direct quantiles for TFT \\
CRPS samples (test/val) & $N=100$ / $N=16$ ($N=4$ for Traffic/Wikipedia validation) \\
\bottomrule
\end{tabular}
\end{table}
\noindent {Dataset-specific architecture.} For Traffic, we use an additional direct $x_t$ connection at the output head.

\subsection{Capacity-Matched Diffusion Baseline}
\label{app:capacity-matched}

TSDiff-Wide increases hidden width from 64 to 128, yielding 498-502k parameters versus 408-506k for Control-TS, including the external forecaster, which accounts for approximately 7-24\% of the total depending on the configuration. It uses the same data protocol, validation-CRPS checkpoint selection, and 100-path evaluation. We report mean and standard deviation over three independent runs with different random seeds. Table~\ref{tab:capacity_matched_with_cm} gives the complete-system comparison.

\begin{table}[H]
\centering
\caption{Complete-system capacity comparison: mean$\pm$SD. Control-TS includes the backbone, Controller, and external forecaster; its results are from Table~\ref{tab:main_results}. Best mean CRPS is \textbf{bold}.}
\vspace{0.55\baselineskip}
\label{tab:capacity_matched_with_cm}
\small
\setlength{\tabcolsep}{5pt}
\begin{tabular}{l|cc|cc}
\toprule
& \multicolumn{2}{c|}{TSDiff-Wide} & \multicolumn{2}{c}{Control-TS} \\
\cmidrule(lr){2-3}\cmidrule(lr){4-5}
Dataset & Params [k] & CRPS & Params [k] & CRPS \\
\midrule
    Electricity  & 501.4 & 0.048$\pm$0.000 & 505.6 & \textbf{0.047$\pm$0.000} \\
    Exchange     & 501.5 & 0.012$\pm$0.000 & 493.1 & \textbf{0.008$\pm$0.000} \\
    KDDCup       & 501.4 & 0.345$\pm$0.002 & 504.2 & \textbf{0.249$\pm$0.001} \\
    M4-Hourly    & 498.8 & 0.035$\pm$0.000 & 407.9 & \textbf{0.022$\pm$0.000} \\
    Solar        & 501.4 & 0.409$\pm$0.001 & 412.4 & \textbf{0.341$\pm$0.000} \\
    Traffic      & 501.4 & \textbf{0.095$\pm$0.000} & 411.0 & 0.097$\pm$0.000 \\
    UberTLC      & 501.4 & \textbf{0.167$\pm$0.000} & 504.2 & 0.169$\pm$0.000 \\
    Wikipedia    & 498.4 & 0.222$\pm$0.000 & 489.4 & \textbf{0.200$\pm$0.000} \\
\bottomrule
\end{tabular}
\end{table}

DeepAR-Large and TFT-Large match the complete-system budget; Table~\ref{tab:capacity_matched_forecasters} reports three independent training runs, each evaluated once.

\begin{table}[H]
\centering
\caption{Capacity-comparable explicit forecasters: mean$\pm$SD. Control-TS results are from Table~\ref{tab:main_results}. Best values are \textbf{bold}.}
\label{tab:capacity_matched_forecasters}
\vspace{0.4em}
\small
\setlength{\tabcolsep}{4.2pt}
\resizebox{\textwidth}{!}{%
\begin{tabular}{lcc cc cc}
\toprule
& \multicolumn{2}{c}{DeepAR-Large}
& \multicolumn{2}{c}{TFT-Large}
& \multicolumn{2}{c}{Control-TS} \\
\cmidrule(lr){2-3}\cmidrule(lr){4-5}\cmidrule(lr){6-7}
Dataset & Params [k] & CRPS & Params [k] & CRPS & Params [k] & CRPS \\
\midrule
Electricity & 502.7 & \textbf{0.047$\pm$0.001} & 493.5 & 0.048$\pm$0.000 & 505.6 & \textbf{0.047$\pm$0.000} \\
Exchange & 496.5 & 0.022$\pm$0.006 & 500.3 & 0.010$\pm$0.001 & 493.1 & \textbf{0.008$\pm$0.000} \\
KDDCup & 502.7 & 0.380$\pm$0.019 & 493.5 & 0.354$\pm$0.044 & 504.2 & \textbf{0.249$\pm$0.001} \\
M4-Hourly & 502.7 & 0.047$\pm$0.006 & 493.5 & 0.030$\pm$0.003 & 407.9 & \textbf{0.022$\pm$0.000} \\
Solar & 502.7 & 0.371$\pm$0.020 & 493.5 & 0.552$\pm$0.045 & 412.4 & \textbf{0.341$\pm$0.000} \\
Traffic & 502.7 & \textbf{0.092$\pm$0.005} & 493.5 & 0.099$\pm$0.010 & 411.0 & 0.097$\pm$0.000 \\
UberTLC & 502.7 & 0.164$\pm$0.005 & 493.5 & \textbf{0.161$\pm$0.002} & 504.2 & 0.169$\pm$0.000 \\
Wikipedia & 497.0 & 0.227$\pm$0.006 & 500.3 & 0.205$\pm$0.001 & 489.4 & \textbf{0.200$\pm$0.000} \\
\bottomrule
\end{tabular}%
}
\end{table}

\subsection{Interaction-Study Summary}
\label{app:interaction-summary}

Table~\ref{tab:interaction-summary} counts datasets on which each mechanism has strictly lower CRPS than the TSDiff backbone in Table~\ref{tab:interaction_levels}.

\begin{table}[H]
\centering
\caption{Datasets on which each interaction mechanism improves the backbone.}
\label{tab:interaction-summary}
\small
\begin{tabular}{lc}
\toprule
Mechanism & Better than the backbone \\
\midrule
Ensembling & 4/8 \\
Input concatenation & 4/8 \\
FiLM & 4/8 \\
1-block residual & 3/8 \\
2-block residual & 3/8 \\
Control-TS & \textbf{8/8} \\
\bottomrule
\end{tabular}
\end{table}

\subsection{Frozen and Fine-Tuned Conditioning}
\label{app:frozen-conditioning}

Table~\ref{tab:frozen-conditioning} separates fine-tuned and frozen-backbone variants of input concatenation and FiLM. The fine-tuned entries are those reported in Table~\ref{tab:interaction_levels}.

As visible in the results, holding the backbone fixed is a less favorable setting for concatenation and FiLM. Residual control, including the one-block and two-block variants as well as full-depth Control-TS, trains a separate pathway and keeps the backbone fixed. Input concatenation and FiLM have no such pathway. Concatenation appends the quantile trajectories to the noisy denoiser input, so they are read by the backbone's input projection. FiLM maps them to scale and shift applied to intermediate backbone features. In both cases the quantile interface is carried by backbone parameters. Freezing those parameters removes the weights that would adapt to the conditioning, which is consistent with the weaker frozen-backbone results. The comparison reported in the main text therefore uses the fine-tuned denoiser. Even with the backbone held fixed, some runs improve on the backbone's results.

Output ensembling does not appear in Table~\ref{tab:frozen-conditioning}. It does not train either forecaster: the diffusion backbone and the external model stay fixed, and only a mixture weight is selected by validation CRPS.

\begin{table}[H]
\centering
\caption{CRPS for fine-tuned and frozen-backbone input concatenation and FiLM. Fine-tuned entries match Table~\ref{tab:interaction_levels}. Best values are \textbf{bold}.}
\label{tab:frozen-conditioning}
\vspace{0.4em}
\small
\setlength{\tabcolsep}{5pt}
\begin{tabular}{lcccc}
\toprule
& \multicolumn{2}{c}{Input concatenation} & \multicolumn{2}{c}{FiLM} \\
\cmidrule(lr){2-3}\cmidrule(lr){4-5}
Dataset & Fine-tuned & Frozen backbone & Fine-tuned & Frozen backbone \\
\midrule
Electricity & \textbf{0.062$\pm$0.000} & 0.065$\pm$0.000 & \textbf{0.054$\pm$0.000} & 0.060$\pm$0.000 \\
Exchange & \textbf{0.007$\pm$0.000} & 0.009$\pm$0.000 & \textbf{0.008$\pm$0.000} & 0.010$\pm$0.000 \\
KDDCup & \textbf{0.289$\pm$0.002} & 0.332$\pm$0.001 & \textbf{0.268$\pm$0.005} & 0.292$\pm$0.001 \\
M4-Hourly & \textbf{0.027$\pm$0.001} & 0.034$\pm$0.001 & \textbf{0.026$\pm$0.001} & 0.040$\pm$0.001 \\
Solar & \textbf{0.308$\pm$0.004} & 0.390$\pm$0.005 & \textbf{0.360$\pm$0.002} & 0.423$\pm$0.003 \\
Traffic & \textbf{0.117$\pm$0.000} & 0.120$\pm$0.001 & \textbf{0.116$\pm$0.001} & 0.119$\pm$0.000 \\
UberTLC & \textbf{0.187$\pm$0.001} & 0.190$\pm$0.000 & \textbf{0.184$\pm$0.001} & 0.189$\pm$0.000 \\
Wikipedia & \textbf{0.241$\pm$0.001} & 0.246$\pm$0.001 & \textbf{0.235$\pm$0.001} & 0.245$\pm$0.001 \\
\bottomrule
\end{tabular}
\end{table}

\subsection{Additional Generative Baseline Protocols}
\label{app:additional-baselines}

We adapt the authors' K$^2$VAE and CN-Diff implementations \citep{wu2025k2vae,rishi2025cndiff} to one target channel, using Control-TS's GluonTS splits, context/horizon lengths, rolling test windows, and per-window mean-absolute scaling. These are task-matched univariate evaluations, not reproductions of the original multivariate benchmarks.

Both use Adam and 100 batches/epoch. Checkpoints and hyperparameters are selected by validation CRPS, evaluated each epoch with 16 paths (four for Traffic/Wikipedia); testing uses 100 paths.

\textbf{K$^2$VAE.} We retain reversible normalization and use 50 epochs, gradient clipping 0.5, and batch size 32 for Electricity, 64 otherwise. We compare a globally learned Koopman operator with a combined formulation that adds a local least-squares estimate, using learning rates $\{10^{-3},10^{-4}\}$. Validation selects the global-only variant throughout and $10^{-3}$ for M4-Hourly/Wikipedia, $10^{-4}$ otherwise. Some combined-operator runs are unstable with one target channel. KDDCup/M4-Hourly/UberTLC use the released Electricity architecture. Patch lengths are 10 for Exchange/Wikipedia and eight otherwise, dividing context and horizon lengths.

\textbf{CN-Diff.} We retain last-observation centering and use up to 100 epochs, batch size 64, learning rate $10^{-3}$, and early-stopping patience 10. The validation-CRPS plateau scheduler uses factor 0.5 and patience five. Searching widths $\{64,256\}$ selects 256 for M4-Hourly/Solar, 64 otherwise. Sampling uses 100 steps with the released quadratic schedule. Changes from upstream are the 100-batch epoch cap, plateau scheduler, and 100 rather than 10 test paths. We also initialize the previously uninitialized temporal-layer weights and biases in the nonlinear transformation $T_\phi$, using the same routine as for its feature layer.
\subsection{TSDiff Refinement Protocol}
\label{app:tsdiff-refine}
We apply the official TSDiff refinement procedure \citep{kollovieh2023tsdiff} to the frozen backbone from which Control-TS is trained, initialized by the DeepAR checkpoints that provide its DeepAR controls. For each test window, 100 DeepAR samples are refined for 20 iterations by likelihood maximization (ML-Q; learning rate 0.1) or Langevin sampling (LMC-Q; step size 0.1, noise scale 0.1), both with the quantile regularizer. The diffusion step used as the density surrogate is chosen by the original heuristic on one training batch. Both variants start from the same forecasts, no model parameters are updated, and hyperparameters follow the released configuration. Evaluation uses the same test windows, 100 forecast paths, and CRPS computation as Control-TS. We report mean$\pm$SD over three sampling seeds. TFT is not used because the procedure refines sample paths, whereas TFT outputs quantiles.

\subsection{Per-Variant Forecasting Results}
\label{app:variant_results}

For each dataset and quantile set (Q1, Q3, or Q5), Table~\ref{tab:main_results} reports the test CRPS of Control-TS using the external forecaster selected by validation CRPS. These validation-selected configurations need not achieve the lowest test CRPS in Table~\ref{tab:variant_results}. Here we provide the complete variant-level CRPS for all six configurations: Control-TS-Q1, Control-TS-Q3, and Control-TS-Q5, each instantiated with either DeepAR or TFT as the external probabilistic forecaster (Section~\ref{subsec:control-signal}, Appendix~\ref{app:control-models}). All other components are kept fixed to the configurations described in Appendix~\ref{app:hparams}, and the TSDiff baseline is included for reference.

Table~\ref{tab:variant_results} shows how test CRPS varies with the external forecaster and quantile granularity. TFT-based variants attain the lowest test CRPS on Electricity, Exchange, Solar, and UberTLC, and DeepAR-based variants on KDDCup, M4-Hourly, and Traffic; the two forecasters tie on Wikipedia. On Solar, TFT-Q5 achieves a test CRPS 0.001 lower than DeepAR-Q5, but validation CRPS selects DeepAR-Q5. On KDDCup, validation selects TFT-Q3 (0.249) although DeepAR-Q5 attains 0.245. On every dataset, a multi-quantile variant (Q3 or Q5) attains the lowest test CRPS, with Q1 tying on Exchange, consistent with the granularity discussion in Section~\ref{subsec:quantile-richness}.

\begin{table}[H]
\centering
\caption{Per-variant CRPS on the eight benchmark datasets for each external forecaster (rows) and quantile variant Q1/Q3/Q5 (Section~\ref{subsec:control-signal}), with the TSDiff baseline for reference. The best entry per column is shown in \textbf{bold} and the second-best is \underline{underlined}. Tied entries are marked jointly.}
\vspace{0.55\baselineskip}
\label{tab:variant_results}
\small
\setlength{\tabcolsep}{4pt}
\resizebox{\textwidth}{!}{%
\begin{tabular}{@{}llcccccccc@{}}
\toprule
External forecaster & Variant & Electricity & Exchange & KDDCup & M4-Hourly & Solar & Traffic & UberTLC & Wikipedia \\
\midrule
\multicolumn{2}{@{}l}{Baseline}
& 0.049 & 0.012 & 0.312 & 0.034 & 0.402 & 0.099 & 0.172 & 0.211 \\
\midrule
\multirow{3}{*}{DeepAR}
& Q1 & 0.054 & \underline{0.009} & 0.303 & \underline{0.025} & 0.419 & \underline{0.098} & 0.177 & 0.203 \\
& Q3 & 0.057 & \underline{0.009} & 0.267 & 0.026 & 0.397 & \textbf{0.097} & 0.171 & 0.205 \\
& Q5 & 0.056 & \underline{0.009} & \textbf{0.245} & \textbf{0.022} & \underline{0.341} & 0.100 & 0.173 & \textbf{0.200} \\
\midrule
\multirow{3}{*}{TFT}
& Q1 & 0.049 & \textbf{0.008} & 0.282 & 0.028 & 0.408 & 0.107 & \underline{0.170} & 0.204 \\
& Q3 & \underline{0.048} & \textbf{0.008} & \underline{0.249} & 0.026 & 0.392 & 0.102 & \textbf{0.169} & \underline{0.202} \\
& Q5 & \textbf{0.047} & \textbf{0.008} & 0.256 & \underline{0.025} & \textbf{0.340} & 0.101 & 0.172 & \textbf{0.200} \\
\bottomrule
\end{tabular}%
}
\end{table}

\subsection{Observed Context as Control}
\label{app:context-control}

Both sides use the same full-depth residual architecture. The Control-TS entries are the Q1, Q3, and Q5 rows of Table~\ref{tab:main_results}. Context only repeats the observed context once, three times, or five times, matching that channel count. It includes observed historical lag channels, zeros all forecast-horizon input channels, and does not load or call an external forecaster. The Solar Q1 result appears to be an outlier, but Control-TS with Q5 still achieves lower CRPS.

\begin{table}[H]
\centering
\caption{\textbf{Observed context as control.} CRPS (mean$\pm$SD) using the same full-depth residual architecture. Control-TS values are from Table~\ref{tab:main_results}. The better value in each pair is \textbf{bold}, including ties.}
\label{tab:context-control}
\small
\setlength{\tabcolsep}{3.5pt}
\resizebox{\textwidth}{!}{%
\begin{tabular}{lcccccc}
\toprule
& \multicolumn{2}{c}{Q1} & \multicolumn{2}{c}{Q3} & \multicolumn{2}{c}{Q5} \\
\cmidrule(lr){2-3}\cmidrule(lr){4-5}\cmidrule(lr){6-7}
Dataset & Control-TS & Context $\times 1$ & Control-TS & Context $\times 3$ & Control-TS & Context $\times 5$ \\
\midrule
Electricity & 0.049$\pm$0.000 & \textbf{0.047$\pm$0.000} & \textbf{0.048$\pm$0.000} & 0.049$\pm$0.000 & \textbf{0.047$\pm$0.000} & 0.048$\pm$0.000 \\
Exchange & \textbf{0.008$\pm$0.000} & 0.010$\pm$0.000 & \textbf{0.008$\pm$0.000} & 0.011$\pm$0.000 & \textbf{0.008$\pm$0.000} & 0.012$\pm$0.000 \\
KDDCup & \textbf{0.282$\pm$0.001} & 0.311$\pm$0.001 & \textbf{0.249$\pm$0.001} & 0.319$\pm$0.001 & \textbf{0.256$\pm$0.001} & 0.282$\pm$0.001 \\
M4-Hourly & \textbf{0.025$\pm$0.000} & 0.045$\pm$0.000 & \textbf{0.026$\pm$0.000} & 0.033$\pm$0.000 & \textbf{0.022$\pm$0.000} & 0.030$\pm$0.000 \\
Solar & 0.408$\pm$0.000 & \textbf{0.375$\pm$0.000} & \textbf{0.392$\pm$0.000} & 0.449$\pm$0.000 & \textbf{0.341$\pm$0.000} & 0.440$\pm$0.001 \\
Traffic & \textbf{0.098$\pm$0.000} & 0.100$\pm$0.000 & \textbf{0.097$\pm$0.000} & 0.099$\pm$0.000 & 0.100$\pm$0.000 & \textbf{0.099$\pm$0.000} \\
UberTLC & \textbf{0.170$\pm$0.000} & 0.171$\pm$0.001 & \textbf{0.169$\pm$0.000} & 0.176$\pm$0.001 & \textbf{0.172$\pm$0.000} & 0.180$\pm$0.001 \\
Wikipedia & \textbf{0.203$\pm$0.000} & 0.212$\pm$0.000 & \textbf{0.202$\pm$0.000} & 0.209$\pm$0.000 & \textbf{0.200$\pm$0.000} & 0.206$\pm$0.000 \\
\bottomrule
\end{tabular}%
}
\end{table}

\subsection{Architectural Ablations}
\label{app:arch-ablations}

This subsection evaluates three design choices in Control-TS: \textbf{(i)} zero-initialized channel projection operators, \textbf{(ii)} the initialization of the Controller residual blocks, and \textbf{(iii)} depth-matched residual injection through all three Controller blocks. Unless otherwise stated, all ablated variants use the same frozen diffusion backbone, training protocol, external forecaster family, and quantile granularity as the corresponding Control-TS configuration selected by validation CRPS for each dataset (Table~\ref{tab:main_results}). We report CRPS on the full test set.

\paragraph{Projection initialization.}
Table~\ref{tab:arch_ablation_init} compares the default Control-TS model against a full-depth variant without zero-initialized channel projections. Removing zero-initialization degrades performance on all eight datasets and is particularly harmful on Solar, Wikipedia, M4-Hourly, and KDDCup, consistent with the role of zero-initialized projections in preserving the pretrained backbone behavior at the start of training and introducing control gradually.

\begin{table}[H]
\centering
\caption{Projection-initialization ablation for full-depth Control-TS. Best values are \textbf{bold}.}
\label{tab:arch_ablation_init}
\small
\setlength{\tabcolsep}{5pt}
\begin{tabular}{lcc}
\toprule
Dataset & Control-TS & No zero-init \\
\midrule
Electricity & \textbf{0.047} & 0.051 \\
Exchange    & \textbf{0.008} & 0.009 \\
KDDCup      & \textbf{0.249} & 0.302 \\
M4-Hourly   & \textbf{0.022} & 0.027 \\
Solar       & \textbf{0.341} & 0.433 \\
Traffic     & \textbf{0.097} & 0.113 \\
UberTLC     & \textbf{0.169} & 0.184 \\
Wikipedia    & \textbf{0.200} & 0.250 \\
\bottomrule
\end{tabular}
\end{table}

\paragraph{Controller initialization.}
For completeness, we compare random initialization of the controller residual blocks (Control-TS) with initialization from the corresponding pretrained backbone weights, keeping the remaining setup unchanged. Table~\ref{tab:controller_initialization} reports the comparison.

\begin{table}[H]
\centering
\caption{Controller-initialization ablation for full-depth Control-TS (CRPS; lower is better). Control-TS initializes its controller residual blocks randomly. Pretrained initialization instead starts each controller block from a copy of the corresponding residual block of the pretrained diffusion backbone; the copies are trained while the backbone remains frozen, and all other settings are unchanged.}
\label{tab:controller_initialization}
\small
\setlength{\tabcolsep}{5pt}
\begin{tabular}{lcc}
\toprule
Dataset & Pretrained initialization & Control-TS \\
\midrule
Electricity & 0.050$\pm$0.000 & 0.047$\pm$0.000 \\
Exchange    & 0.009$\pm$0.000 & 0.008$\pm$0.000 \\
KDDCup      & 0.305$\pm$0.000 & 0.249$\pm$0.001 \\
M4-Hourly   & 0.030$\pm$0.000 & 0.022$\pm$0.000 \\
Solar       & 0.401$\pm$0.000 & 0.341$\pm$0.000 \\
Traffic     & 0.107$\pm$0.001 & 0.097$\pm$0.000 \\
UberTLC     & 0.175$\pm$0.000 & 0.169$\pm$0.000 \\
Wikipedia    & 0.220$\pm$0.000 & 0.200$\pm$0.000 \\
\bottomrule
\end{tabular}
\end{table}

\paragraph{Controller-depth ablations.}
Table~\ref{tab:arch_ablation_depth} studies whether residual corrections need to be injected throughout the full backbone depth. We compare the default three-block Controller against shallower variants that inject residual corrections only through block 0, or through blocks 0 and 1. The default three-block model has the lowest or tied-lowest displayed CRPS on all eight datasets; 
Exchange, the two-block variant ties it at 0.008 after rounding. Among the shallow variants, using two Controller blocks is generally better than using one, indicating that control benefits from being injected beyond the earliest layer. Overall, these results support the use of depth-matched residual control.

\begin{table}[H]
\centering
\caption{Controller-depth ablations. \textbf{bold} identifies the full-depth reference, which is lowest or tied-lowest at displayed precision.}
\label{tab:arch_ablation_depth}
\small
\setlength{\tabcolsep}{5pt}
\begin{tabular}{lccc}
\toprule
Dataset & 1 block & 2 blocks & Control-TS \\
\midrule
Electricity & 0.052 & 0.052 & \textbf{0.047} \\
Exchange    & 0.009 & 0.008 & \textbf{0.008} \\
KDDCup      & 0.286 & 0.285 & \textbf{0.249} \\
M4-Hourly   & 0.032 & 0.026 & \textbf{0.022} \\
Solar       & 0.444 & 0.415 & \textbf{0.341} \\
Traffic     & 0.125 & 0.118 & \textbf{0.097} \\
UberTLC     & 0.203 & 0.193 & \textbf{0.169} \\
Wikipedia    & 0.248 & 0.245 & \textbf{0.200} \\
\bottomrule
\end{tabular}
\end{table}

\subsection{Robustness to Degraded Control Signals}
\label{app:robustness}

To complement the representative robustness results shown in the main text, we report here the full evaluation across all eight datasets and all tested perturbation levels, including the more extreme setting $\sigma=0.1$. As in the main-text experiment, we perturb the quantile-based control signal at inference time by adding Gaussian noise in normalized control space, while keeping both the trained Controller and the frozen diffusion backbone fixed. We then evaluate CRPS on the full test set for each dataset.

Table~\ref{tab:gaussian_noise_ablation} reports the resulting mean CRPS values under increasing noise levels, together with the unguided diffusion-backbone baseline. Entries better than the backbone are \textbf{bold}; ties are \underline{underlined}. The degradation is clearly dataset-dependent. KDDCup is the most robust: Control-TS remains better than the backbone at every tested noise level, including the stress-test setting $\sigma=0.1$. Wikipedia is similarly stable, and ties the backbone even at the strongest perturbation. Solar also degrades gradually and remains better than the backbone up to $\sigma=0.05$, crossing above it only at $\sigma=0.1$. In contrast, Electricity and M4 are more sensitive: both remain improved under mild perturbations ($\sigma=0.01$ and $\sigma=0.02$), but fall below the backbone at $\sigma=0.05$ and deteriorate further at $\sigma=0.1$. Traffic and UberTLC show a similar but milder pattern, staying competitive under small perturbations before losing their advantage under stronger corruption. Exchange is the most fragile among the eight datasets: although Control-TS remains better than the backbone at $\sigma=0.01$, the gain disappears already at $\sigma=0.02$ and degrades substantially thereafter. Taken together, these results show that Control-TS is generally tolerant to mild noise in the control signal, but that stronger degradation can negate its benefit in a distinctly dataset-dependent manner.

\begin{figure}[t]
    \centering
    \includegraphics[width=0.95\textwidth]{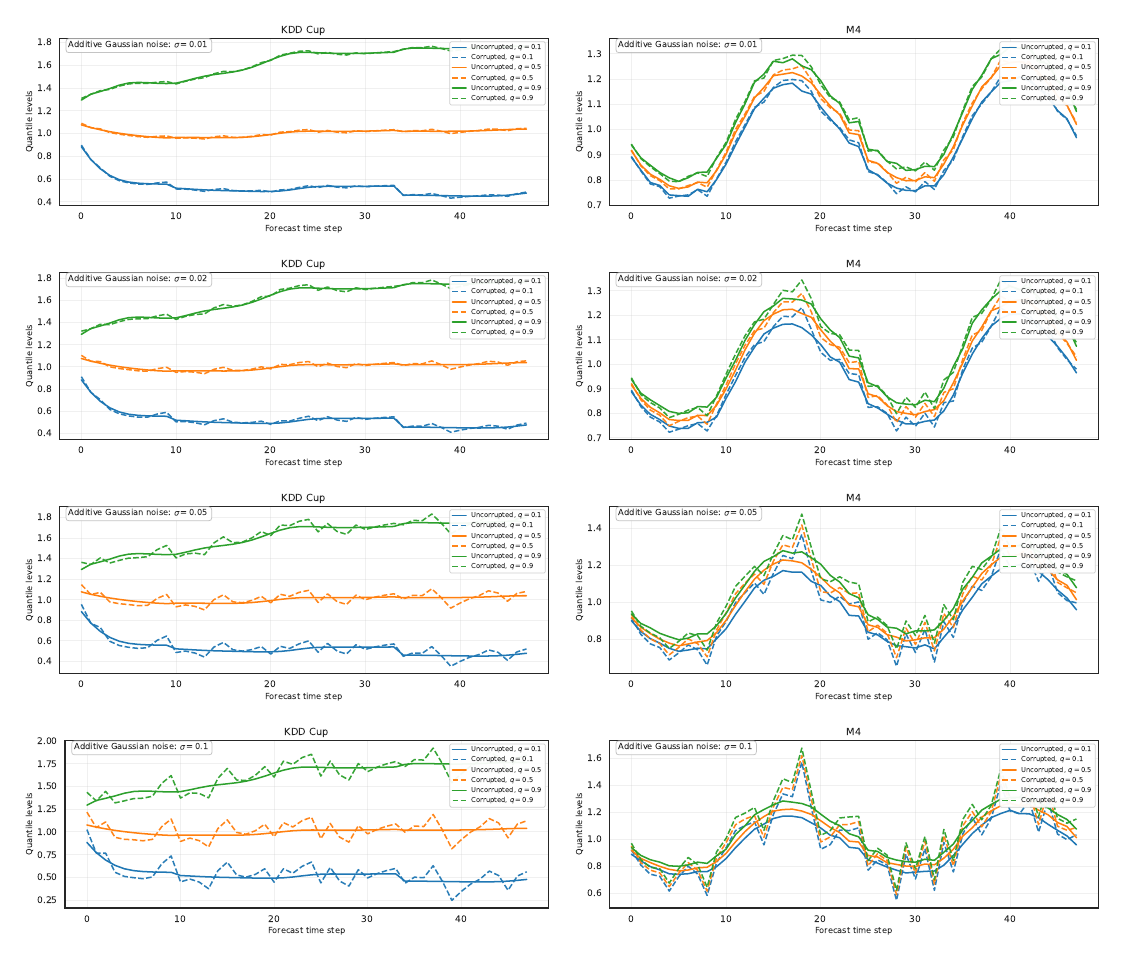}
    \caption{Examples of corrupted quantile-control signals under additive Gaussian perturbations in normalized control space. Solid curves show the original quantile trajectories and dashed curves show the perturbed controls for representative datasets and increasing noise levels.}
    \label{fig:merged_noise_figure}
\end{figure}

Figure~\ref{fig:merged_noise_figure} visualizes the effect of Gaussian perturbations on the quantile-based control signal for two representative datasets, KDDCup and M4, across noise levels $\sigma \in \{0.01, 0.02, 0.05, 0.1\}$. In each panel, the corrupted quantile trajectories are shown against their uncorrupted counterparts for $q=0.1$, $0.5$, and $0.9$. As expected, the perturbation is mild at small noise levels and becomes progressively more disruptive as $\sigma$ increases, eventually distorting both the central trajectory and the uncertainty envelope. This visualization complements Table~\ref{tab:gaussian_noise_ablation} by showing directly how Gaussian degradation weakens the uncertainty scaffold provided to the Controller.

\begin{table}[H]
\centering
\caption{CRPS under Gaussian degradation of the control signal at inference time. Noise is added in normalized control space. Entries better than the unguided diffusion-backbone baseline are shown in \textbf{bold}, while ties are \underline{underlined}.}
\label{tab:gaussian_noise_ablation}
\small
\setlength{\tabcolsep}{5.0pt}
\begin{tabular}{lcccccc}
\toprule
Dataset & Clean & $\sigma=0.01$ & $\sigma=0.02$ & $\sigma=0.05$ & $\sigma=0.1$ & Backbone \\
\midrule
Electricity & \textbf{0.047} & \textbf{0.048} & 0.049 & 0.058 & 0.080 & 0.049 \\
Exchange    & \textbf{0.008} & \textbf{0.010} & 0.014 & 0.031 & 0.065 & 0.012 \\
KDDCup      & \textbf{0.249} & \textbf{0.249} & \textbf{0.251} & \textbf{0.252} & \textbf{0.261} & 0.312 \\
M4          & \textbf{0.022} & \textbf{0.023} & \textbf{0.024} & 0.037 & 0.053 & 0.034 \\
Solar       & \textbf{0.341} & \textbf{0.374} & \textbf{0.377} & \textbf{0.389} & 0.424 & 0.402 \\
Traffic     & \textbf{0.097} & \textbf{0.098} & \underline{0.099} & 0.108 & 0.131 & 0.099 \\
UberTLC     & \textbf{0.169} & \textbf{0.170} & \textbf{0.169} & 0.174 & 0.182 & 0.172 \\
Wikipedia    & \textbf{0.200} & \textbf{0.201} & \textbf{0.201} & \textbf{0.203} & \underline{0.211} & 0.211 \\
\bottomrule
\end{tabular}
\end{table}

\subsection{Failure Cases under Misaligned Control Signals}
\label{app:bad_control_failure}

\begin{figure}[t]
\centering
\includegraphics[width=0.88\textwidth,height=0.99\textheight,keepaspectratio]{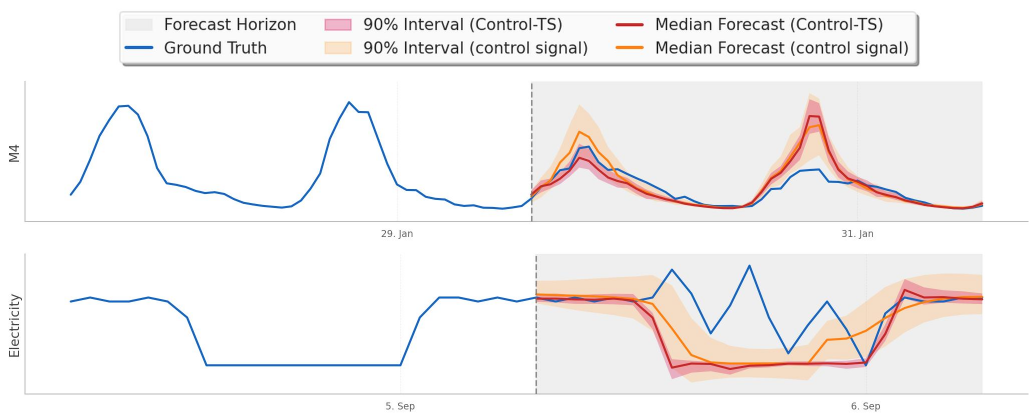}
\caption{Failure cases of Control-TS: misaligned quantile scaffolds where the forecast follows the unreliable signal.}
\label{fig:bad_control_failure}
\end{figure}

Beyond the controlled Gaussian perturbations of Appendix~\ref{app:robustness}, we also examine naturally occurring cases in which the external forecaster's quantile scaffold is misaligned with the realized future. While the diffusion backbone typically buffers Control-TS against such misalignment (Section~\ref{subsec:control_signal_quality}, Figure~\ref{fig:buffered_control}), Figure~\ref{fig:bad_control_failure} shows representative cases where the controller does follow a misaligned scaffold and the resulting forecast drifts toward the unreliable signal. These cases are uncommon in our experiments but are reported for transparency and to bound the practical reliability of residual control under in-the-wild miscalibration of the external forecaster.

\subsection{Computational Overhead: Full Accounting}
\label{app:compute}

This appendix expands the analysis of Section~\ref{subsec:compute-overhead}. We describe the measurement protocol, the parameter budget of Control-TS relative to the frozen diffusion backbone, the structural reasons inference cost shows modest growth while training cost is more pronounced, and report the per-dataset values for all eight benchmarks corresponding to Table~\ref{tab:compute_full_8}.

\textbf{Protocol.} For each of the eight benchmarks we load checkpoints for both the frozen backbone and the trained Control-TS, build one batch of size $B{=}16$ from the standard test split, and measure on a single NVIDIA RTX 2080 Ti: \textbf{(i)} end-to-end sampling wall-clock, FLOPs, and peak VRAM for one forecast call ($T_{\mathrm{diff}}{=}100$ reverse-diffusion steps with quantile observation self-guidance, see Appendix~\ref{app:diffusion}); and \textbf{(ii)} wall-clock, FLOPs, and peak VRAM for one denoising training step (forward pass, backward pass, and optimizer update). Wall-clock is averaged over $5$ inference runs after $5$ warm-up runs and over $100$ training-step runs after $10$ warm-up runs. FLOPs are obtained via \texttt{torch.profiler} with FLOP attribution enabled. Peak VRAM is the maximum allocated GPU memory at a given moment during the same call.

\textbf{Parameter accounting.} A trained Control-TS model comprises three components: \textbf{(a)} the frozen backbone, whose parameters $\theta_B$ are held fixed throughout Controller training; \textbf{(b)} a trainable Controller branch, consisting of a residual stack matching the backbone's architecture together with a Controller input projection $E_{\theta_C}$ acting on the control signal $c$; and \textbf{(c)} zero-initialized $1\times 1$ channel projection operators $\{Z_{\theta_C}^{(k)}\}_{k=1}^{K}$ that gate residual injection at each backbone depth (see Section~\ref{subsec:architecture} and Appendix~\ref{app:architecture-details}). At training, only the Controller branch and its zero-initialized projections receive gradient updates; the diffusion backbone is held fixed. The Controller branch is therefore approximately the same size as the backbone but slightly smaller, because it does not include the backbone's input/output projections, time-embedding head, or lag-feature head. Per-dataset parameter counts are reported in Table~\ref{tab:compute_params}.

\begin{table}[H]
\centering
\caption{Parameter budget of Control-TS per dataset. The diffusion backbone is held fixed during Controller training; only the Controller branch receives gradient updates. Counts in thousands of parameters (k); ratio is Controller~/~Backbone.}
\vspace{0.55\baselineskip}
\label{tab:compute_params}
\small
\begin{tabular}{lccc}
\toprule
Dataset & Backbone (frozen) [k] & Controller (trainable) [k] & Controller~/~Backbone \\
\midrule
Electricity & 193.4 & 191.4 & 0.99 \\
Exchange    & 193.4 & 190.1 & 0.98 \\
KDDCup      & 193.4 & 190.0 & 0.98 \\
M4          & 192.1 & 188.2 & 0.98 \\
Solar       & 193.4 & 191.4 & 0.99 \\
Traffic     & 193.4 & 190.0 & 0.98 \\
UberTLC     & 193.4 & 190.0 & 0.98 \\
Wikipedia   & 191.9 & 188.0 & 0.98 \\
\bottomrule
\end{tabular}
\end{table}

\textbf{Why inference cost modestly grows.} At each reverse-diffusion timestep $t$, the sampler performs two forward passes through the denoising network: an unconditional reverse-kernel mean (Eq.~\ref{eq:ddpm-mean}), evaluated with the controlled denoiser $\hat{\epsilon}_{\theta}(x_t,t\mid c)$ for Control-TS, and an observation self-guidance score $-\nabla_{x_t} E$ (Eq.~\ref{eq:obs-score}), which differentiates through the backbone-only Tweedie one-step estimate $\tilde{x}_{0,B}$ (Eq.~\ref{eq:x0hat}, evaluated with the unmodified backbone) with respect to the noisy sample $x_t$. For both models, the self-guidance pass is evaluated through the unmodified diffusion backbone and accounts for roughly two-thirds of per-step compute in our profiler splits. Only the unconditional reverse-kernel mean differs across models. Because the Controller mirrors only the residual block stack of the backbone (and not its input/output projections, time-embedding head, or lag-feature head), this slice grows by $\sim\!1.2\times$-$1.7\times$, leading to a per-step total ratio of $\sim\!1.07\times$-$1.25\times$ over $T_{\mathrm{diff}}{=}100$ steps--matching the inference FLOPs columns of Table~\ref{tab:compute_full_8}.

\textbf{Why inference VRAM gap is small.} At inference, all diffusion parameters are frozen (\texttt{requires\_grad}=False), so the autograd tape used to evaluate the self-guidance score retains only the activations needed for the input gradient $\partial E/\partial x_t$, not weight gradients. Since this score is the architectural bottleneck of the reverse-diffusion step and is computed through the unmodified backbone, peak inference VRAM is set by the same self-guidance tape for both models, and only a small Controller forward-only contribution from the unconditional reverse-kernel mean is added on top. Peak inference VRAM therefore grows by less than $7\%$ across all datasets.

\textbf{Why training cost grows more.} Training requires gradients through both the controlled denoising network and the noise-prediction loss, retaining activations through both backbone (frozen) and Controller. Across datasets, peak training VRAM rises by a consistent $64\%$ to $66\%$ as the Controller's activation tape is added to the existing backbone tape. Training-step wall-clock stays below $15\%$ on six of eight datasets and below $66\%$ on all of them, so the per-step training budget never doubles. To isolate the Controller's intrinsic forward$+$backward cost, the timing window excludes the per-step regeneration of control quantiles (constructed once outside the timing window): in training-mode, control quantiles are cached upstream rather than regenerated per step, so including the GluonTS \texttt{predict} path inside the window would inflate per-step time with framework overhead unrelated to the Control-TS forward/backward. Per-step training FLOPs from the profiler are reported alongside.

\textbf{Full per-dataset table.} Table~\ref{tab:compute_full_8} reports all six measurements for all eight benchmarks. Inference overhead is small and bounded across datasets ($17\%$ to $61\%$ wall-clock, $7\%$ to $25\%$ FLOPs, $2\%$ to $7\%$ peak VRAM), while training overhead is consistent in VRAM ($64\%$ to $66\%$) and bounded in wall-clock--under $20\%$ on six of eight benchmarks and under $66\%$ on all of them.

\begin{table}[htb]
\centering
\caption{Computational overhead of Control-TS relative to the frozen backbone for all eight benchmarks ($B{=}16$, $T_{\mathrm{diff}}{=}100$, with quantile observation self-guidance). Each $\Delta$ column is the relative increase (\%) of Control-TS over the diffusion backbone; GFLOPs columns report ${\color{tsdiffcolor}\text{Backbone}}\,/\,{\color{controltscolor}\text{Control-TS}}$.}
\vspace{0.55\baselineskip}
\label{tab:compute_full_8}
\small
\setlength{\tabcolsep}{3.2pt}
\resizebox{\textwidth}{!}{%
\begin{tabular}{lccc ccc}
\toprule
& \multicolumn{3}{c}{Inference (sampling)} & \multicolumn{3}{c}{Training (one step)} \\
\cmidrule(lr){2-4}\cmidrule(lr){5-7}
Dataset & $\Delta$Time (\%) & GFLOPs & $\Delta$VRAM (\%) & $\Delta$Time (\%) & GFLOPs & $\Delta$VRAM (\%) \\
\midrule
Electricity & $+19.1$ & {\color{tsdiffcolor}21.5}\,/\,{\color{controltscolor}26.8} & $+6.1$ & $+13.5$ & {\color{tsdiffcolor}0.20}\,/\,{\color{controltscolor}0.24} & $+66.2$ \\
Exchange    & $+61.1$ & {\color{tsdiffcolor}24.2}\,/\,{\color{controltscolor}28.7} & $+4.0$ & $+65.6$ & {\color{tsdiffcolor}0.23}\,/\,{\color{controltscolor}0.24} & $+66.0$ \\
KDDCup      & $+17.1$ & {\color{tsdiffcolor}21.5}\,/\,{\color{controltscolor}25.2} & $+5.9$ & $+13.2$ & {\color{tsdiffcolor}0.20}\,/\,{\color{controltscolor}0.21} & $+64.8$ \\
M4          & $+39.2$ & {\color{tsdiffcolor}17.0}\,/\,{\color{controltscolor}18.2} & $+3.5$ & $+6.5$  & {\color{tsdiffcolor}0.17}\,/\,{\color{controltscolor}0.14} & $+63.9$ \\
Solar       & $+21.4$ & {\color{tsdiffcolor}21.5}\,/\,{\color{controltscolor}26.4} & $+4.5$ & $+14.9$ & {\color{tsdiffcolor}0.20}\,/\,{\color{controltscolor}0.24} & $+65.0$ \\
Traffic     & $+21.0$ & {\color{tsdiffcolor}21.5}\,/\,{\color{controltscolor}24.8} & $+4.2$ & $+10.6$ & {\color{tsdiffcolor}0.20}\,/\,{\color{controltscolor}0.21} & $+64.3$ \\
UberTLC     & $+30.3$ & {\color{tsdiffcolor}21.5}\,/\,{\color{controltscolor}25.2} & $+5.9$ & $+14.7$ & {\color{tsdiffcolor}0.20}\,/\,{\color{controltscolor}0.21} & $+64.8$ \\
Wikipedia   & $+31.5$ & {\color{tsdiffcolor}18.4}\,/\,{\color{controltscolor}20.1} & $+2.4$ & $+47.3$ & {\color{tsdiffcolor}0.18}\,/\,{\color{controltscolor}0.16} & $+65.7$ \\
\bottomrule
\end{tabular}%
}
\end{table}

\textbf{Take-away.} Self-guidance dominates inference cost and is shared between the frozen diffusion backbone and Control-TS, so the Controller adds only a bounded slice of compute and memory at sampling time. Training is where the Controller is fully paid for, but only as a one-time amortized cost during Controller fitting, after which a Control-TS model is queried at inference at near-baseline cost. This profile makes residual control attractive whenever a pretrained diffusion forecaster is already deployed: the Controller can be fit and shipped alongside the backbone with only a marginal sampling-time penalty.

\subsection{CLPS Details: THTR Construction and Linear Evaluator}
\label{app:clps-details}
We compute CLPS under a Train-on-Hybrid, Test-on-Real (THTR) evaluation using $N_{\mathrm{hyb}}=20{,}000$ hybrid samples, where ``hybrid'' refers to real observed contexts paired with synthetic forecast horizons sampled from the model. We hold the training contexts fixed across TSDiff and Control-TS, so only the generated futures differ, with no test leakage into the hybrid data. We sample $N_{\mathrm{hyb}}$ real contexts $\mathbf{y}_{\mathrm{ctx}}$ from the training set and generate one synthetic target horizon per context by running model inference conditioned on $\mathbf{y}_{\mathrm{ctx}}$ (and its corresponding control signal $c$), yielding hybrid pairs $(\mathbf{y}_{\mathrm{ctx}}, \hat{\mathbf{y}}_{\mathrm{tar}})$. We then fit a closed-form ridge forecaster on these hybrid pairs and evaluate it on the real test set using CRPS. Because the linear evaluator is deterministic, it produces a single forecast trajectory for each test context. To use the same CRPS evaluation pipeline as the generative models, we replicate this trajectory across the 100 sampled paths used for CRPS computation; these draws are therefore identical and introduce no additional Monte Carlo variability. For large $N_{\mathrm{hyb}}$, drawing one synthetic target per context yields $N_{\mathrm{hyb}}$ i.i.d.\ hybrid samples from the joint distribution
\[
p_{\mathrm{hyb}}(\mathbf{y}_{\mathrm{ctx}}, \hat{\mathbf{y}}_{\mathrm{tar}})
=
p_{\mathrm{data}}(\mathbf{y}_{\mathrm{ctx}})
\, p_{\mathrm{gen}}(\hat{\mathbf{y}}_{\mathrm{tar}} \mid \mathbf{y}_{\mathrm{ctx}}),
\]
and Monte Carlo estimation error for statistics of these pairs decays as $O\!\left(\frac{1}{\sqrt{N_{\mathrm{hyb}}}}\right)$ \citep{mcbook}, which is adequate for the stability of the closed-form linear evaluator used by CLPS.

\subsection{Combined Additional-Backbone Scope Study}
\label{app:backbone-transfer}

Paired CSDI--DeepAR and MG-TSD--DeepVAR evaluations use matched windows and 100 paths (Table~\ref{tab:scope-paired}). CSDI uses CRPS; multivariate MG-TSD uses CRPS-sum, scoring the sum across variates.

\textbf{Integration.} Both keep native forecasting, with frozen backbones and external forecasters. Quantile controls are the Controller's input; they are not concatenated into CSDI's side information or into MG-TSD's recurrent history encoders. %
The Controller adds a zero-initialized projection at the input of each corresponding backbone residual block. CSDI uses four 64-channel residual layers, eight attention heads, and no feature-axis attention (univariate). MG-TSD uses eight 8-channel residual blocks shared across granularities 1 and 4; its two-layer 128-unit recurrent encoders remain frozen backbone modules, and control is applied at each autoregressive step. Q5 controls use $\{0.1,0.25,0.5,0.75,0.9\}$ from 100 external paths, with predicted horizon quantiles appended to the repeated scaled context and aligned to the forecast origin and backbone scale. For MG-TSD's coarser granularity, the horizon control is the quantile of DeepVAR paths after averaging each coarse time block, not the average of per-step quantiles.

\begin{table}[H]
\centering
\caption{Paired test-score reductions (\%) from Control-TS, using 100 forecast paths. Positive values indicate improvement; negative values indicate regression.}
\label{tab:scope-paired}
\label{tab:scope-multivariate}
\small
\setlength{\tabcolsep}{8pt}
\begin{tabular}{@{}lrr@{}}
\toprule
& \multicolumn{1}{c}{CSDI: univariate} & \multicolumn{1}{c}{MG-TSD: multivariate} \\
\cmidrule(lr){2-2}\cmidrule(lr){3-3}
Dataset & CRPS & CRPS-sum \\
\midrule
Electricity & $+7.90$ & $+6.35$ \\
Exchange & $+9.98$ & $-4.16$ \\
KDDCup & $+3.90$ & $-7.23$ \\
M4-Hourly & $+9.73$ & $+3.98$ \\
Solar & $+0.34$ & $+38.27$ \\
Traffic & $+11.05$ & $+1.91$ \\
UberTLC & $+1.58$ & $-2.29$ \\
Wikipedia & $+2.67$ & $+8.48$ \\
\bottomrule
\end{tabular}
\end{table}

\textbf{CSDI training.} Backbone/Controller: Adam $10^{-3}$, batch size 64, up to 100 epochs of 100 batches, early-stopping patience 10, plateau factor 0.5/patience five, and validation-CRPS selection. DeepAR: fitting-only data, 15 epochs of 100 batches, two 40-unit layers, dropout 0.1, batch size 32, and validation negative-log-likelihood selection. Sampling uses all 50 ancestral steps, with quadratic noise from $10^{-4}$ to $0.5$.

\textbf{MG-TSD training.} Both use 30 epochs of 100 updates; effective batch size is 128 for Solar, 32 otherwise. The baseline uses OneCycle (peak 0.01); the Controller uses fixed-rate Adam $10^{-3}$, clipping 0.5. Validation CRPS-sum at epoch 1 and every five epochs selects checkpoints. Sampling: 100 steps, linear noise ending at 0.1, sharing ratios $\{1,0.6\}$, loss weights $\{0.9,0.1\}$. DeepVAR: two 128-unit LSTM layers, dropout 0.1, rank-10 Gaussian output (eight for Exchange), 200 epochs of 100 updates, Adam $10^{-3}$, marginal Gaussian CRPS; validation selects fixed-rate/plateau candidates.

\textbf{Splits and evaluation.} Fitting excludes validation/test targets. CSDI uses reference context/horizon lengths and seven validation windows for Electricity/Solar/Traffic, five for Exchange/Wikipedia, one otherwise. MG-TSD reserves three validation horizons; context equals horizon. Dimensions/horizons follow Table~\ref{tab:dataset_overview}'s training-series counts/prediction lengths. Wikipedia fitting ends before the first test origin. MG-TSD panels use calendar time, except relative-origin alignment for KDDCup/M4-Hourly/UberTLC.

\clearpage

\subsection{Additional Qualitative Examples}
\label{app:qualitative-examples}
\begin{figure}[H]
\centering
\includegraphics[width=0.98\textwidth,height=0.9\textheight,keepaspectratio]{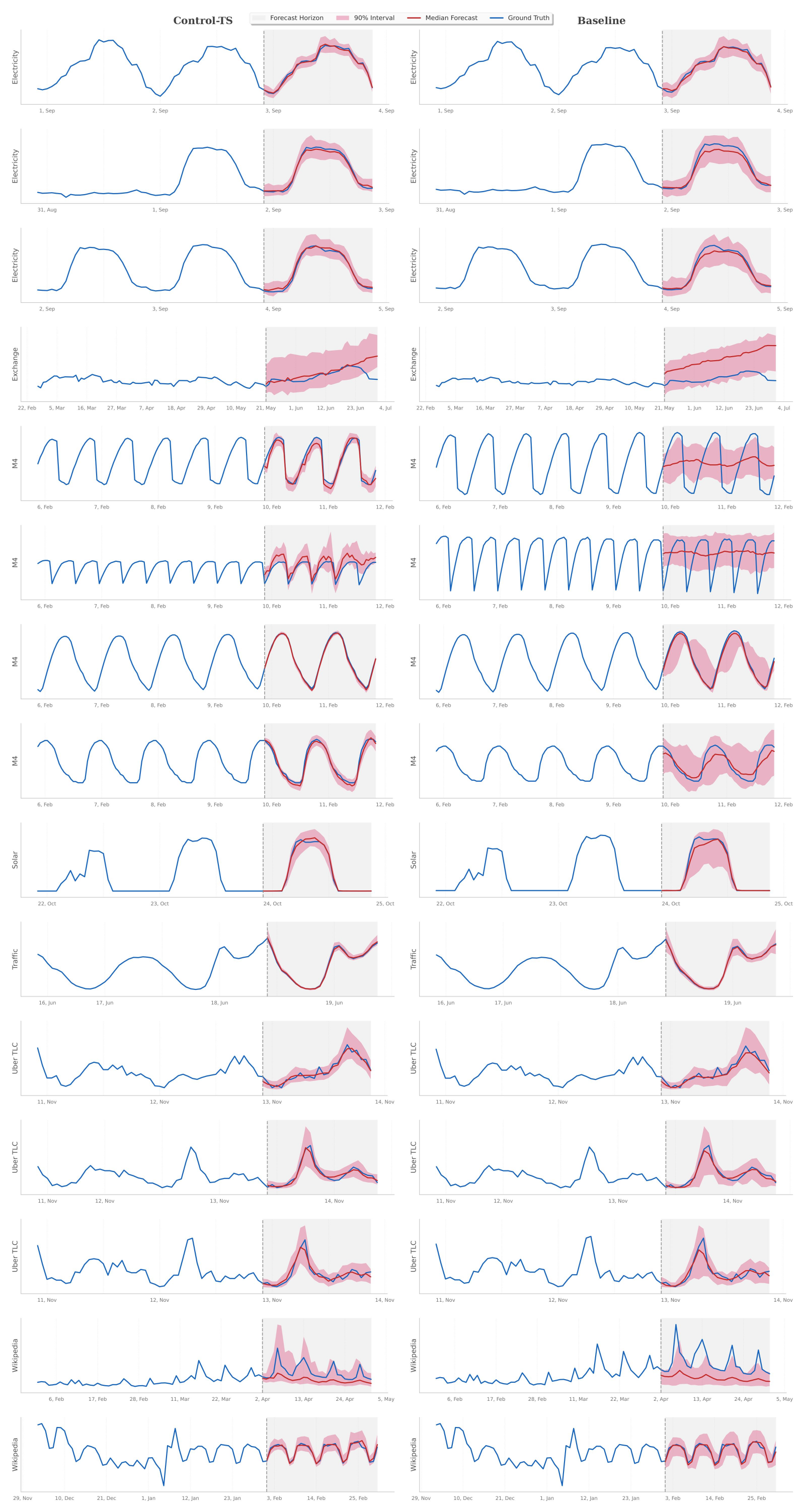}
\caption{Additional qualitative comparisons between Control-TS and the TSDiff baseline across selected benchmark series.}
\label{fig:qualitative_examples}
\end{figure}

\clearpage
\begin{figure}[H]
\centering
\includegraphics[width=0.88\textwidth,height=0.99\textheight,keepaspectratio]{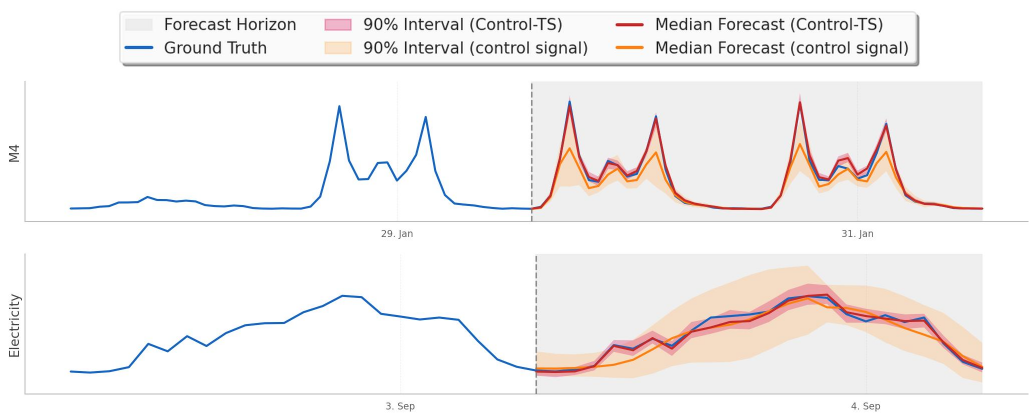}
\caption{Examples with accurate quantile scaffolds: Control-TS tightens its predictive distribution around the realized future.}
\label{fig:good_quantiles_goodcontrolts}
\end{figure}

\end{document}